\documentclass{article}

\usepackage{natbib}

\usepackage[utf8]{inputenc} 
\usepackage[T1]{fontenc}    
\usepackage{hyperref}       
\usepackage{url} 
\usepackage{bbm}
\usepackage{booktabs}       
\usepackage{amsfonts}       
\usepackage{nicefrac}       
\usepackage{microtype}      
\usepackage{xcolor}         
\usepackage{amsmath, amssymb, amsthm}
\usepackage{graphicx}
\usepackage{verbatim}
\usepackage{zref-xr}
\usepackage{xspace}
\makeatletter
\newcommand*{\addFileDependency}[1]{
  \typeout{(#1)}
  \@addtofilelist{#1}
  \IfFileExists{#1}{}{\typeout{No file #1.}}
}
\makeatother

\newcommand*{\myexternaldocument}[1]{%
    \zexternaldocument{#1}%
    \addFileDependency{#1.tex}%
    \addFileDependency{#1.aux}%
}

\myexternaldocument{appendix}

\newtheorem{definition}{Defnition}
\newtheorem{assumption}{Assumption}
\newtheorem{lemma}{Lemma}
\newtheorem{theorem}{Theorem}
\newtheorem{remark}{Remark}

\newcommand{\sj}[1]{\textcolor{blue}{[SJ: #1]}}
\newcommand{\al}[1]{\textcolor{green}{[AL: #1]}}
\newcommand{\nisha}[1]{\textcolor{violet}{[nisha: #1]}}

\newcommand{\E}{\mathbb E}
\newcommand{\Exp}{\mathbb{E}}

\usepackage[margin=1.1in]{geometry}
\renewcommand{\paragraph}[1]{\vspace{1.0mm}\noindent\textbf{#1}}
\title{On the generalization of learning algorithms that do not converge}

\author{%
  Nisha Chandramoorthy\\
  Institute for Data, Systems and Society\\
  Massachusetts Institute of Technology\\
  Cambridge, MA 02139\\
  \texttt{nishac@mit.edu} \\
  \and
  Andreas Loukas \\
  Prescient Design \\
  Genentech, Roche \\
  \texttt{andreas.loukas@roche.com} \\
  \and
  Khashayar Gatmiry \\
  Electrical Engineering and Computer Science\\
  Massachusetts Institute of Technology\\
  Cambridge, MA 02139\\
  \texttt{gatmiry@mit.edu} \\
  \and
  Stefanie Jegelka \\
  Electrical Engineering and Computer Science\\
  Massachusetts Institute of Technology\\
  Cambridge, MA 02139\\
  \texttt{stefje@mit.edu}
}

\begin{document}

\maketitle

\begin{abstract}
Generalization analyses of deep learning typically assume that the training converges to a fixed point. But, recent results indicate that in practice, the weights of deep neural networks optimized with stochastic gradient descent often oscillate indefinitely. 
To reduce this discrepancy between theory and practice, this paper focuses on the generalization of neural networks whose training dynamics do not necessarily converge to fixed points. 
Our main contribution is to propose a notion of \textit{statistical algorithmic stability} (SAS) that extends classical algorithmic stability to non-convergent algorithms and to study its connection to generalization. This ergodic-theoretic approach leads to new insights when compared to the traditional optimization and learning theory perspectives. 
We prove that the stability of the time-asymptotic behavior of a learning algorithm relates to its generalization and empirically demonstrate how loss dynamics can provide clues to generalization performance. Our findings provide evidence that networks that ``train stably generalize better'' even when the training continues indefinitely and the weights do not converge. 
\end{abstract}

\section{Introduction}

It is common practice that when the training loss of a neural networks converges close to some value, the learning algorithm---typically some variant of Stochastic Gradient Descent (SGD)---is terminated. 
Perhaps surprisingly, recent works indicate that the network function at termination time is typically not a fixed point of the learning algorithm: if run longer, 
the gradient norm does not vanish and the learning algorithm outputs functions with significantly different parameters \citep{cohen, suvrit, lobacheva2021periodic}. This observation stands in contrast to common generalization analyses that assume convergence of the training algorithm to a fixed point. 
It also raises the question of when (and why) non-convergent learning algorithms should be expected to generalize.
%

Standard approaches for obtaining generalization bounds find ways to bound the complexity measure of the function class expressible by a neural network in a data-independent e.g., ~\citep{vapnik1999overview, bartlett2017spectrally,neyshabur2018pac,golowich2018size,JMLR:v20:17-612,arora2018stronger} or data-dependent manner, e.g., ~\citep{von2004distance,xu2012robustness,sokolic2017robust}. 
Nevertheless, even with recent technical improvements, by and large these approaches do not explicitly account for the relationship between generalization and the dynamics of the learning algorithm.


A different approach to generalization analysis yields algorithm-dependent bounds by connecting the generalization performance of the trained network to the stability of the training to data perturbations \citep{bousquet2002stability, feldmanVondrak, kuzborskij2018data,algstab, zhang2021stability, rakhlin2006applications}. A key claim of these works has been that networks that are trained fast generalize better. This claim, though intuitively meaningful, hinges on the premise of a convergent algorithm which deviates from the observed behavior of deep neural network training. Further work has been done for settings where the algorithm provably converges to a fixed point, such as two-layer overparameterized networks, deep linear networks, certain matrix factorizations, etc.\xspace\citep{frei, allen-zhu, soudry2018implicit, aroraNTK,gunasekar, matrixFact}. In a more general setting, \citet{hardtRecht} prove algorithmic stability-based generalization bounds that worsen with increasing training time, whereas~\citet{loukas2021what} connect stable training dynamics near convergence under Dropout with good generalization. For algorithms that do not converge (close) to a fixed point, the above analyses result in vacuous bounds. 

\textbf{Contributions.}
We start with the supposition that robustness of the learning algorithm's \textit{fixed points} to small changes in the training set is \textit{not} the mechanism underlying the good generalization properties of deep networks. This is because, as is commonly known and many recent works indirectly remark upon \citep{cohen, sra1, suvrit, baysian1}, training can be \emph{linearly unstable}: 
even if assuming the presence of isolated local minima that the learning algorithm converges to, the weights encountered in training are not robust to small perturbations. Small perturbations can accumulate over time leading to completely different orbits in weight space. Yet, despite unstable dynamics, the generalization properties of two completely different fixed points (e.g., obtained from two different random initializations or due to a different order of SGD updates) are typically comparable.

Our first contribution entails proposing a generalization of stability that applies to non-converging algorithms.
%
%
To achieve this, we depart from the typical optimization perspective of analyzing the generalization properties of minima \citep{keskar2016large, ge2015escaping, sagun2016eigenvalues}. Instead, we study the generalization performance of the learning algorithm \emph{on average} (see section \ref{subsec:setting}). Adopting a statistical viewpoint, in Section~\ref{sec:stochasticStability} we define a new notion of {\em statistical algorithmic stability} that measures the stability of this average performance to perturbations of the training data, as opposed to stability of individual orbits in the weight space. Section~\ref{sec:stochasticStability} then proves upper bounds for the average generalization error based on our new notion of statistical stability.

We then move on to consider how statistical stability connects to the behavior of the loss function.  
%
%
%
To this end, in Section \ref{sec:predict} we provide conditions such 
that the spectral gap of a Markov operator associated with the dynamics of the loss function is predictive of statistical algorithmic stability and hence of generalization. Empirically, we estimate this spectral gap based on the rate of convergence of ergodic averages of the loss function. We show via experiments that our estimate correlates with generalization for image classification under corruption.


\subsection{Related work}

 Our work builds on previous analyses of the stability of learning algorithms that converge~\citep{bousquet2002stability, feldmanVondrak, kuzborskij2018data,algstab, zhang2021stability}. Below, we also briefly discuss connections (beyond the literature on stability) with previous analyses on the dynamics of learning algorithms. 

\paragraph{Convergence and generalization of SGD.}  Exponential convergence rates to global minima are known for SGD in the smooth and convex setting, with decaying learning rates and small constant learning rates (e.g., \citep{moulines, ward}). In the training of deep neural networks, which is the focus of this paper, the loss function is non-convex and is typically not well-approximated locally by convex functions \citep{belkin1}. However, in the overparameterized regime, a Polyak-Lojasiewicz-type inequality that is automatically satisfied when the loss is smooth and convex can be shown to hold in the non-convex setting as well \citep{liu2022loss, belkin1}, allowing convergence to a global minimum. In certain non-convex problems, convergence to local minima and even global minima has been proved, assuming a strict saddle property \citep{ge2015escaping} or local convexity \citep{pmlr-v80-kleinberg18a}. \citet{raginsky2017non} show a favorable convergence for a slightly different version of SGD by approximating the dynamics with a suitable continuous time Langevin diffusion.  A number of previous works~\citep{safran2016quality,freeman2016topology,li2017convergence,nguyen2018loss} have also provided arguments that SGD converges to a solution that generalizes given suitable initialization and significant overparameterization. A few works have studied the nonlinear dynamics of training \citep{kongTao} and provided insights into generalization in the overparameterized regime \citep{dogra2020optimizing, saxe, ADVANI2020428}.   
Another line of work considers the mean field limit of SGD dynamics to show generalization \citep{mei2019mean, gabrie2018entropy, chen2020mean}.
The ergodic theoretic perspective we adopt here deviates from the optimization perspectives but relates to the mean field perspective in that we implicitly consider (see section \ref{sec:predict}) the evolution of probability distributions on weight space.

\paragraph{Flat minima.} In the generalization literature, flat minima correspond to large connected regions with low empirical error in weight space. Flat minima have been argued to be related to the complexity of the resulting hypothesis and, hence, can imply good generalization~\cite{10.1162/neco.1997.9.1.1}. It has also been shown that SGD converges more frequently to flat minima when the batch size is small or the learning rate and the number of iterations are suitably adjusted~\citep{keskar2016large,hoffer2017train,jastrzkebski2017three,smith2018bayesian,zhang2018theory}. Some works consider the flat/sharp dichotomy an oversimplification~\citep{dinh2017sharp,sagun2017empirical,he2019asymmetric}, e.g., using a reparameterization argument~\citep{sagun2017empirical}. The eigenvalues of the hessian of the loss affect the local linear stability of optimization orbits. However, given that we adopt the time-asymptotic/statistical picture, we do not explicitly make assumptions on local stability (flatness/sharpness) or the topology of the loss landscape \citep{montanari2020interpolation, venturi2019spurious}. 

\paragraph{SGD as a Bayesian sampler.} The idea of looking at distributions of learners permeates PAC-Bayesian analyses of generalization~\citep{mcallester1999some,dziugaite2017computing,zhou2018nonvacuous,pitas2020dissecting}. In contrast to Bayesian posteriors of the parameters, we study parameter distributions generated by the learning algorithms. Previous empirical work has suggested that SGD operates almost like a Bayesian sampler \citep{mingard2021sgd} but the connection remains to be fully understood.

\section{Local descent algorithms as dynamical systems: a statistical viewpoint}

This section lays out the basic definitions and assumptions of our work. We begin in Section~\ref{subsec:ML_basics} by introducing the learning problem and discussing learning algorithms from a dynamical systems perspective. Section~\ref{subsec:setting} then puts forth the notions of statistical convergence that give rise to our main results.

\subsection{Learning as a dynamical system}
\label{subsec:ML_basics}

We consider the supervised learning setup: a learner is given a training set $S = \{z_1, \ldots, z_n\}$ consisting of $n$ pairs $z_i \equiv (x_i, y_i)$ of inputs $x_i \in \mathbb{R}^d$ and their corresponding labels $y_i \in \mathcal{Y} \subseteq \mathbb{R}$, drawn i.i.d. from a distribution $\mathcal{D}$. A class of parametric models or \emph{hypotheses} $h: \mathbb{R}^d \times \mathbb{W} \to \mathbb{R}$ on a parameter space $\mathbb{W}$ gives possible input-to-output relationships, $h(w, \cdot):\mathbb{R}^d\to \mathbb{R}.$ The learner is given a risk or \emph{loss} function $\ell:(\mathbb{R}^d \times \mathbb{R}) \times \mathbb{W} \to \mathbb{R}^+$ that describes the error of a given hypothesis. Common choices for loss functions are the mean-squared, hinge and cross-entropy loss. 
The learner attempts to minimize the population risk
$ R_S(h(\cdot, w)) = \E_{z \sim \mathcal{D}} \ell(z, w),
$ by minimizing the empirical risk
\begin{align}
    \hat{R}_S(h(\cdot, w)) = L_S (w) = \dfrac{1}{n}\sum_{z_i \in S} \ell(z_i, w).
\end{align}
This minimization is achieved using a local descent algorithm given by iterative updates $\phi_S:\mathbb{W}\to\mathbb{W}$ of the form
\begin{align}
\label{eqn:phi}
    w_{t+1} = \phi_S(w_{t}):= w_t - \eta_t \:  \hat{\nabla} L_S(w_t), 
\end{align}
where $\eta_t$ is the learning rate or step size at time $t.$ 

In the gradient descent (GD) algorithm, $\hat{\nabla} L_S(w) = \nabla L_S(w)$ is the gradient w.r.t.\ $w$, and the iterates of $\phi_S$ represent a deterministic dynamical system. The critical points (including saddle points, local and global minima) of $L_S$ are fixed points of $\phi_S$ (i.e., points $w$ such that $\phi_S w = w$). For stochastic gradient descent (SGD), $\hat{\nabla}L_S(w)$ is a random variable given by the sample mean of gradients $\nabla \ell (\cdot, w)$ over a batch of samples in $S$. In this case, $\phi_S$ is a random dynamical system and critical points of $L_S$ are not necessarily fixed points of $\phi_S$. 

To unify the definition of $\phi_S$ for both GD and SGD, we introduce the random variable $\Xi$ that indicates the choice of batch. Suppose the batch size $m$ is fixed: $m < n$ for SGD and $m = n$ for GD. Denote by $[n]$ the set $\left\{1,2,\cdots, n\right\}$ and let $\Xi_t$ be a collection of $m$ elements from $[n]$ chosen uniformly at random. 
Then, we may write $\phi_S$ as
\begin{align}
    w_{t+1} = \phi_S(w_t) =
    w_t - \eta_t \hat{\nabla }L_S(w_t|\Xi_t)
    = w_t - \dfrac{\eta_t}{m}\sum_{i \in \Xi_t} \nabla\ell(z_i, w_t),   
\end{align}
with $\Xi_t$ being the set $[n]$ at all $t$ for GD.

In our investigation, we opt for simplicity and consider a fixed learning rate $\eta_t = \eta$. More generally, our analysis can be extended to learning rates that asymptotically converge to $\eta$. A rapidly vanishing learning rate can obscure the learning dynamics by enforcing convergence to arbitrary points irrespective of the points' losses. In contrast, learning rates that do not decay to zero can induce interesting transitions between neighborhoods of critical points and the iterates $w_t$ are generally more exploratory of the loss landscape.


%


\begin{figure}
    \centering
    \includegraphics[trim=40cm 0 6cm 0cm,clip,width=0.35\textwidth]{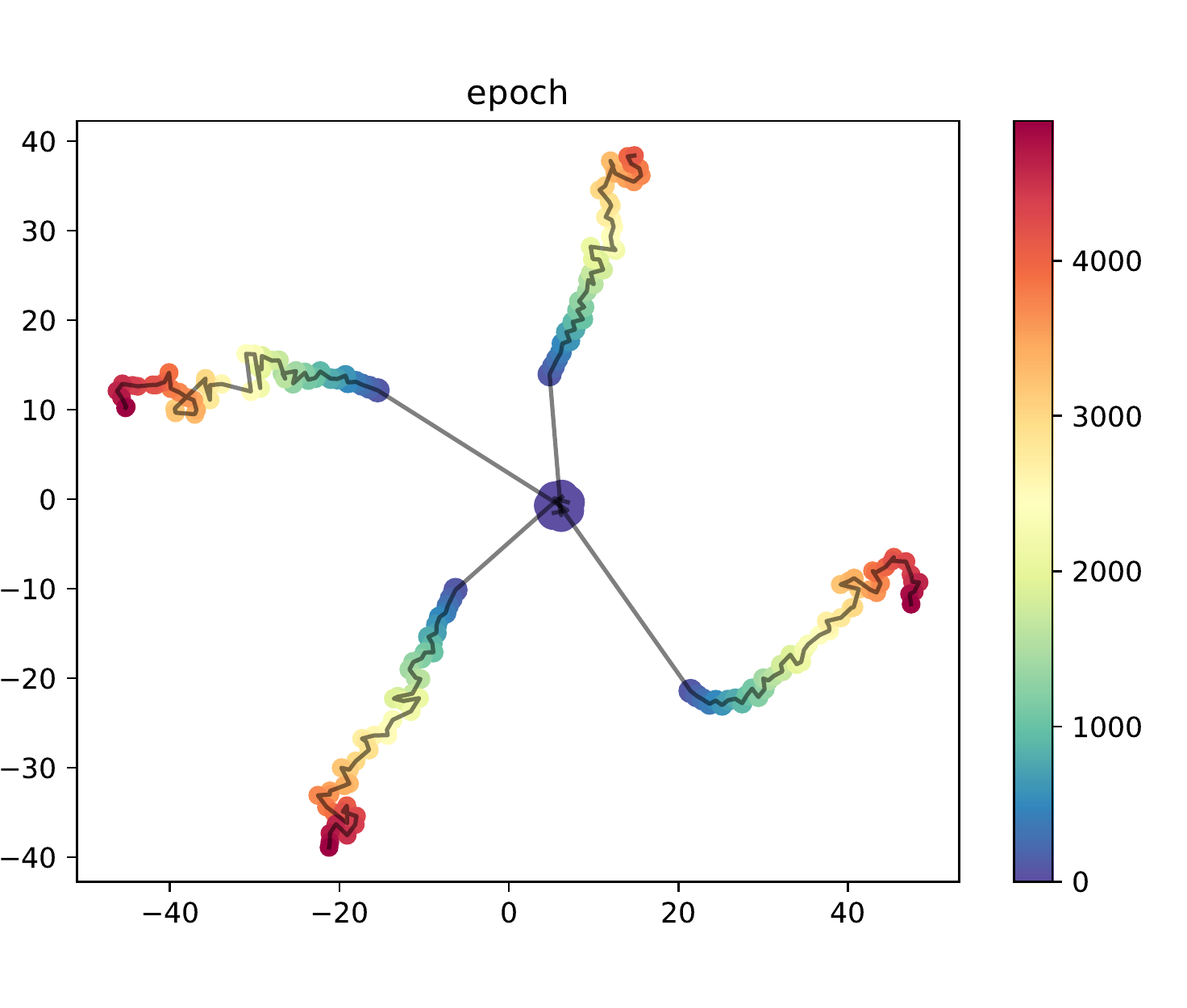}
    ~ 
    \includegraphics[width=0.35\textwidth]{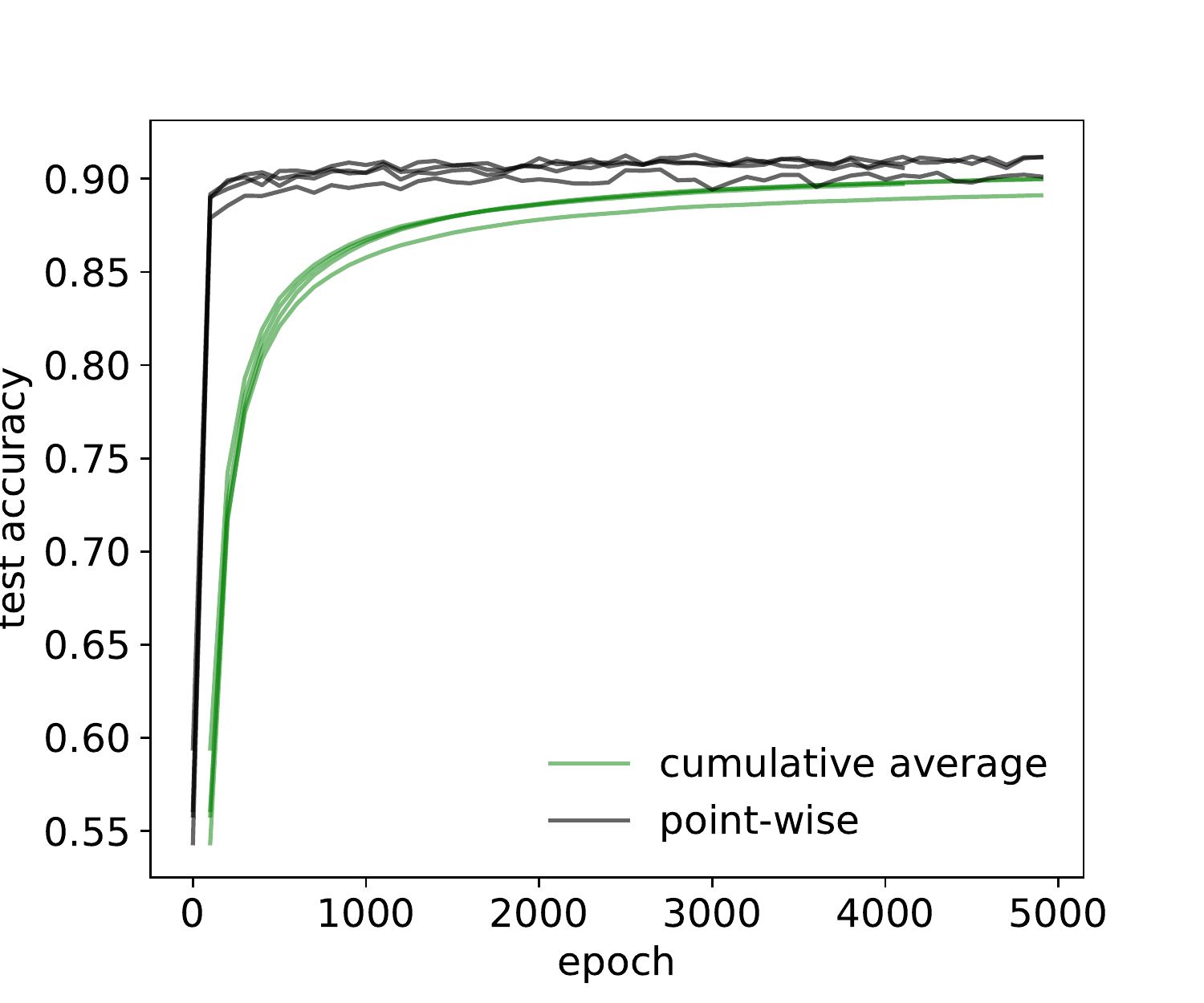}
    \caption{Although the training of neural networks depends on the initialization and does not necessarily converge in the weight space, certain functionals of the learned hypothesis can converge in distribution independently of initial conditions. Left: The orbits of the 2nd layer weights of four different VGG16 models trained on a CIFAR10 dataset using SGD with step size 0.01. We embed the orbits in the plane by performing PCA onto 40 dimensions followed by t-SNE. Colors indicate epoch. Right: Corresponding test accuracy and cumulative average values as a function of epoch.\vspace{-5mm}}
    \label{fig:loss_erg}
\end{figure}

\subsection{Statistical convergence of the learning algorithm}
\label{subsec:setting}

In deep learning, any specific choice of $w_T,$ at some time $T,$ is arbitrary as the loss landscape is generally non-convex and stochastic learning algorithms are the norm. Hence, rather than focusing on a single hypothesis, it is attractive to consider the average generalization error induced by the stochastic dynamical system $\phi_S$. We adopt this \emph{statistical} perspective, rather than focus on a single orbit of $\phi_S.$ Thus, the setting below applies to any asymptotic behavior of orbits of the (typically nonlinear) dynamics $\phi_S$ including fixed points, periodic, quasiperiodic and chaotic orbits.

To analyze the generalization performance without making assumptions on the global stability properties of individual orbits, we make an assumption about the ergodic properties of $\phi_S$  that reasonably matches empirical evidence. To describe the assumption, we give a short primer on the relevant concepts from ergodic theory of dynamical systems. 
%

\paragraph{Invariant measures.} Denote the Borel sigma algebra on $\mathbb{W}$ by $\mathcal{B}(\mathbb{W})$.  A probability measure $\mu_S$ is called \textit{invariant} for $\phi_S$ if $\mu_S(A) = \mu_S\circ \phi_S^{-1}(A)$ for all Borel sets $A \in \mathcal{B}(\mathbb{W})$. Intuitively, for any subset of the weight space, the probability that the dynamical system occupies it does not change. See \citet{liverani2004invariant} for an introduction to invariant measures.

Can we expect the dynamics of learning algorithms to give rise to invariant measures? If $\phi_S$ is a deterministic continuous function on the set $\mathbb{W}$ and $\mathbb{W}$ is compact, the Krylov–Bogolioubov  theorem (see e.g., Theorem 4.1.1 of \citet{katok}) gives the existence of at least one invariant measure for $\phi_S.$ 
On the other hand, when $\phi_S$ is an SGD update, \citet{bach} show the existence of invariant distributions for convex loss functions, whereas \citet{SGDconvergence1} and \citet{MATTINGLY2002185} show the existence of invariant distributions in more general settings under mild assumptions on the gradient noise (in the estimate $\hat{\nabla}L_S).$ Several works have analyzed the asymptotic properties of such invariant distributions, e.g., \citet{phaseTransition} analyze the oscillation of the iterates $w_t$ about a mean that is a critical point, as a {\em phase} separate from the transient phase of convergence to this invariant distribution. \citet{kongTao} study the effect of large learning rates (see also \citep{WangTao} for matrix factorization problems) and multiscale loss functions to show the existence of Gibbs invariant measures in GD.

In general, there are infinitely many invariant measures. A pertinent class of invariant measures in practice is that of ergodic, invariant measures, which allows us to understand the statistical properties of a system through evolution of individual orbits. For an ergodic, invariant measure, sets that are invariant under the dynamics $\phi_S$ (such as a set of periodic points, quasiperiodic or chaotic attractors) are trivial: they have measure 0 or 1.

\paragraph{Ergodicity.} When an invariant measure $\mu_S$ is ergodic for $\phi_S$, time averages converge to ensemble averages according to $\mu_S$.
%
That is, for $\mu_S$-almost every initial state $w_0,$ and for all continuous scalar functions $h$:
\begin{align}
   \label{eq:timeaverage}
    \dfrac{1}{T}\sum_{t=0}^{T-1} h(w_t) \to \E_{w\sim \mu_S} h(w)
\end{align}
and the dependence on initial conditions $w_0$ is forgotten.

Unfortunately, we cannot generally expect the dynamics of the learning algorithm to have a unique ergodic invariant measure on $\mathbb{W}$. Instead, $\phi_S$-invariant sets could be smaller subsets of $\mathbb{W}$ with different ergodic invariant measures supported on them. As a result, starting from two different points chosen Lebesgue almost everywhere (or sampled from any probability density on $\mathbb{W}$), we do not expect time averages to converge to the same limit.  Indeed, as shown in  Figure~\ref{fig:loss_erg}, the weights visited by $\phi_S$ depend on the initial conditions and can vary significantly across different runs. In other words, the limit of infinite time averages described in Equation \eqref{eq:timeaverage} does depend on $w_0.$ 

\paragraph{Dynamics of the hypothesis.} To circumvent the above issues, rather than focusing on the weights directly, we shift perspective and consider the dynamics of the \emph{learned hypothesis} $h(\cdot, w_t)$. Specifically, we posit that the infinite-time averages of the learned hypothesis are ergodic: that is, the time averages of $h(z, w_t)$ converge to the same function irrespective of the initial function $h(z, w_0).$
As a result,  functionals that depend only on $h$ and not explicitly on the weights are also ergodic. Our main assumption is that the time averages of loss functions are ergodic, since they depend only on the network functions:  
\begin{assumption}
\label{ass:ergodicity}
 Given any $S \sim \mathcal{D}^n$, there exists a map $z \to \langle \ell_z\rangle_S$, such that for Lebesgue-a.e. $w_0$ and every $z \in \mathbb{R}^d\times \mathbb{R},$ the following holds:
 \begin{align}
		 \lim_{T\to\infty}\dfrac{1}{T}\sum_{t=0}^{T-1} \ell(z, w_t) = \langle \ell_z\rangle_S \in \mathbb{R},	\label{eq:lossErgodicity}
    \end{align}
  where $\left\{w_t = \phi_S(w_{t-1})\right\}$ is an orbit of $\phi_S.$ 
\end{assumption}
As an intuitive justification for Assumption~\ref{ass:ergodicity}, symmetries may result in several sets of parameters (weights and biases) that represent the same network function $h.$ Hence, assuming that there exists a unique probability on the space of neural network functions that is ergodic is less restrictive and closer to reality than assuming a unique ergodic measure on the parameter space. Furthermore, ergodicity of the network functions is a stronger assumption than our Assumption~\ref{ass:ergodicity} and serves to illustrate one possible sufficient condition for our assumption to hold.

Figure~\ref{fig:loss_erg} provides evidence that Assumption~\ref{ass:ergodicity} can hold in practice. It shows two observables calculated along 4 different orbits of a VGG16 model trained using SGD with momentum (batches of size 128, learning rate of 0.01, momentum of 0.9). On the left, we can see a low-dimensional projection of the second layer weights obtained by projecting the weights onto 40 dimensions using PCA and then employing t-SNE \citep{tsne}, whereas the observable on the right is the test accuracy, which is a functional of the hypothesis function.
Here, the test accuracy can be seen to converge to a distribution independent of the initial conditions, whilst the weight orbits vary significantly across different initializations. This experiment corroborates our hypothesis that time averages of functionals on the loss space converge to distributions independent of the initial condition, even if time averages of a generic observable on the weight space do not.

\begin{remark}{Increasing the learning rate may induce ergodicity.}
{\rm 
Assumption \ref{ass:ergodicity} may hold for all continuous functions for sufficiently large constant learning rates, as long as orbits do not diverge. Intuitively, large learning rates can cause more frequent transitions from the basin of attraction of one local minimum to another. On the other hand, for small learning rates, we may be able to detect the presence of multiple attractors, since different initial conditions lead to different long-time averages. These ideas are illustrated with a toy example in Appendix \ref{sec:exampleDynamicsNonConvex}.}
\end{remark} 

\section{Statistical stability implies generalization}
\label{sec:stochasticStability}



In Section~\ref{subsec:sas}, we generalize algorithmic stability beyond algorithms that converge to fixed points. Section~\ref{subsec:sas_generalization} then proceeds to examine the implications of our definition to generalization error.

\subsection{Statistical algorithmic stability}
\label{subsec:sas}

Classically, the derivations of algorithmic stability-based generalization (see e.g., \citet{hardtRecht}, Chapter 14 of \citet{mohri}, \citet{algstab}) utilize an input perturbation of one element in $S$ by replacing it with a different element from the input distribution $\mathcal{D}$:

%

\begin{definition}[Algorithmic stability, adapted from~\citep{bousquet2002stability}]
\label{def:algorithmicStability}
Consider the weights $w_S^*$ and $w_{S'}^*$ obtained by running the learning algorithm on two training sets $S$, $S'$ sampled from $\mathcal{D}$ that differ by exactly one sample. We say that the learning algorithm is algorithmically stable (AS) with a stability coefficient $\beta \geq 0$ if 
\begin{align} 
    \beta = \sup \left\{ |\ell(z,w_S^*) - \ell(z,w_{S'}^*)|:  z \in \mathbb{R}^d\times \mathbb{R}\right\}.
\label{eq:alg_stability}	
\end{align}
\end{definition}
We refer to this type of perturbation as a \emph{stochastic} perturbation. Algorithmic stability does not directly apply to learning algorithms that do not converge to a fixed point. Hence, next, we extend algorithmic stability to loss \emph{statistics}. The resulting notion of \emph{statistical algorithmic stability (SAS)}
extends  algorithmic stability to algorithms whose loss statistics converge as prescribed by Assumption \ref{ass:ergodicity}, even if the weights do not converge.
\begin{definition}[Statistical algorithmic stability]
	\label{def:algorithmicStatisticalStability}
We say that a learning algorithm with loss statistics denoted by $\langle\ell_z\rangle_S$ is statistically algorithmically stable (SAS) with a stability coefficient $\beta \geq 0$ if \begin{align} 
\beta = \sup \left\{
    |\langle\ell_z\rangle_S - \langle\ell_z\rangle_{S'}|:  z \in \mathbb{R}^d\times \mathbb{R}
    \right\},
\label{eq:stability}	
\end{align}
for all $S$, $S'$ sampled from $\mathcal{D}$ that differ by exactly one point. 
\end{definition}

In the above definition, $\langle\ell_z\rangle_{S'}$ refers to the ergodic average of $\ell(z,\cdot)$ observed along almost every orbit of $\phi_{S'}.$ A higher value of $\beta$ indicates a lower SAS algorithm. The definition of SAS differs from Definition~\ref{def:algorithmicStability}  as we do not assume convergence of the learning algorithm. Nevertheless, when every orbit of $\phi_S$ converges to a fixed point $w_S^*,$ for almost every $S,$ then Definition~\eqref{eq:stability} reduces to the standard notion of algorithmic stability since ergodic averages along almost every orbit converge to $\ell(z, w_S^*).$ 

Crucially, and in line with the observations in Figure~\ref{fig:loss_erg}, we quantify stability based on the statistics of the \emph{loss function} and not the loss function at an ensemble mean or at any one point in the weight space $\mathbb{W}$. In other words, the definition of SAS does not use or provide information about the algorithmic stability of $\E_{w \in \mu_S} w$ for any invariant measure $\mu_S.$


\subsection{Learning theoretic implications}
\label{subsec:sas_generalization}

It is well-known that Definition~\ref{def:algorithmicStability} leads to generalization bounds (see \citep{hardtRecht, algstab} and references therein). Next, we show that the more general statistical algorithmic stability from Definition~\ref{def:algorithmicStatisticalStability} also can imply a notion of generalization. 

To obtain generalization bounds, we first redefine the empirical and population risk to use the loss statistics rather than loss values at fixed points: 
\begin{align}
	\hat{R}_S = \dfrac{1}{n} \sum_{z \in S} \langle \ell_z\rangle_S \quad \text{and} \quad
\label{eq:generalizationRisk2}
	 R_S = \Exp_{z \sim D} \langle \ell_z\rangle_S.
\end{align}
Although the definitions above appear deceivingly similar with the standard ones introduced in Section~\ref{subsec:ML_basics}, the two notions of risk are defined on different spaces, with the standard ones being maps from the \emph{hypothesis space} $\mathcal{H}$ to scalar values and \eqref{eq:generalizationRisk2} being maps from the space of \emph{learning algorithms}.
Using these definitions, the following generalization bound holds: 

\begin{theorem}
\label{thm:sasImpliesGeneralization}
Given a $\beta$-statistically algorithmically stable algorithm $\phi_S$ (see Definition \ref{def:algorithmicStatisticalStability}), for any $\delta \in (0,1),$ with probability greater than $1-\delta$, 
\begin{align}
\label{eq:generalizationBound}
		R_S &\leq \hat{R}_S + \beta +
		2\left(n\beta +L\right)\sqrt{\dfrac{\log(2/\delta)}{2n}},  
\end{align}
where $L := \sup_z \sup_{w\in \mathbb{W}}\left| \ell(z, w)\right|$ is an upper bound on the loss.
\end{theorem}

The proof can be found in Appendix  \ref{sec:stabilityImpliesGeneralization} and is based on applying Mcdiarmid's inequality to the generalization gap. %
Like \citet{bousquet2002stability}'s bound, the stability coefficient needs to grow at most like $\beta \sim {\cal O}(1/\sqrt{n})$ for the generalization bound to be non-vacuous. Note that our proof is also naturally coupled with \citep{algstab, feldmanVondrak}'s improved technique, which provides tighter bounds when $\beta \sim {\cal O}(1/\sqrt{n}).$ (see Appendix \ref{sec:stabilityImpliesGeneralization}).

\section{Dynamical systems interpretation of SAS}
\label{sec:predict}

How can we distinguish between two algorithms with different SAS? Can the algorithm dynamics provide clues to its SAS? These questions are of practical importance because a more (statistically) stable algorithm generalizes better (via Theorem \ref{thm:sasImpliesGeneralization}). Since the SAS coefficient $\beta$ is defined as a supremum over an infinite number of pairs of training sets and inputs, a numerical estimation of it using its definition only gives a rough lower bound and obtaining this lower bound is also computationally expensive (see Figure \ref{fig:generalization} and section \ref{sec:numeric}). Moreover, a numerical lower bound on $\beta$ does not give us a mechanistic understanding of statistical stability, which we seek here.   

Here we develop an operator-theoretic explanation for what makes an algorithm statistically stable. We also derive a rough heuristic -- albeit not a definitive predictor of generalization -- that relates a given training dynamics to its SAS coefficient $\beta$. 

\subsection{Bounding the SAS coefficient}

Before we present our bound, we discuss why we must deviate from the 
trajectory-based approach that is commonly employed to study algorithmic stability. 
We recall that \citet{hardtRecht} base their proofs of classical algorithmic stability (Theorems 3.7 and 3.8 in \citet{hardtRecht}) on the accumulating differences between two orbits corresponding to $S$ and $S',$ whenever the different element is chosen in the mini-batch. The greater the difference in the weights, the greater the difference in the (upper bound on the) loss functions at those weights, in the case of Lipschitz loss. Thus, \citet{hardtRecht} argue that training faster (or stopping earlier) will lead to stabler algorithms. Since SAS is about robustness of loss statistics or \textit{infinitely} long time averages, the analysis \`a la \citet{hardtRecht} generically leads to vacuous bounds for the SAS coefficient. 

Our analysis is based on the realization that an algorithm can be SAS even if two orbits corresponding to $S$ and $S'$ diverge from each other (within $\mathbb{W})$, but lead to similar loss statistics. Thus, to bound the SAS coefficient, a perturbation analysis of  transition operators (global information), rather than a finite-time perturbation analysis of an orbit (local information), is needed. Since SAS only demands robustness on loss space, we consider Markov  transition operators on the loss space ($\subseteq [0,L]$), as opposed to on the weight space ($\mathbb{W}$). In general, at any $z$ and $w_0,$ the loss process, $\{\ell(z,w_t)\}_t$ is not Markovian. But, a family of Markov operators can be associated with the family of loss functions. 


\begin{lemma}{(Markov operators)}
\label{lemma:markov}
Let $\mu_S$ be an ergodic, invariant measure for $\phi_S.$ Assume that a loss function $\ell(z, \cdot):\mathbb{W}\to I_z \subseteq [0,L]$ is such that 
the pushforward, $\nu_S^z:\mathcal{B}(I_z)\to \mathbb{R}^+,$ of $\mu_S$ on the loss space is well-defined. Here,  $\mathcal{B}(I_z)$ is the Borel sigma algebra on $I_z$ and $\nu_S^z = \ell(z,\cdot)_\sharp \mu_S,$ where $\sharp$ refers to the pushforward operation. Then, there exists a uniformly ergodic Markov operator $\mathcal{P}_{\mu_S}^z$ on the space of probability measures on $I_z,$ with its invariant measure being $\nu_S^z.$
\end{lemma}
This lemma (see Appendix \ref{sec:correlation} for a proof) parallels \citet{chekroun2014}'s Theorem A, but in a quite different setting, without using existence of a unique physical measure on the phase space ($\mathbb{W}$). 
Combined with Assumption \ref{ass:ergodicity}, perturbation results on the Markov operators, $\mathcal{P}_{\mu_S}^z,$ defined by Lemma \ref{lemma:markov} lead us to SAS. Since $\mathcal{P}_{\mu_S}^z$ is uniformly ergodic, there exist $\lambda_z \in (0, 1)$ and $C > 0$ such that for all $\xi \in I_z,$ in the Wasserstein metric,
\begin{align}
\label{eq:uniformErgodicity}
    \|\mathcal{P}_{\mu_S}^{z^t}\delta_\xi - \nu_S^z\|_{W} \leq C \lambda_z^t. 
\end{align}

Let $\lambda := \sup_z \lambda_z$ be the rate of mixing associated with the family of operators $\mathcal{P}_{\mu_S}^z.$ We leverage the perturbation theory of Markov operators to study the effect of stochastic perturbations on $\nu_S^z,$ and subsequently, SAS. 
To see the connection with SAS, we recall that, by Assumption \ref{ass:ergodicity}, for any choice of $\mu_S,$ we have that $\langle\ell_z\rangle_S = \mathbb{E}_{\xi \sim \nu_S^z} \xi.  $ Thus, given any pair of $\mu_S$ and $\mu_{S'},$ we have that 
\begin{align}
\label{eq:connectionWithSAS}
  |\langle\ell_z\rangle_S - \langle\ell_z\rangle_{S'}|  = |\mathbb{E}_{\xi \sim \nu_S^z} \xi  - \mathbb{E}_{\xi \sim \nu_{S'}^z} \xi| \leq \|\nu_S^z - \nu_{S'}^z\|_{W},
\end{align}
where $\|\cdot - *\|_W$ is the Wasserstein metric (see Appendix \ref{sec:correlation} for details).  Thus, a uniform upper bound on $\|\nu_S^z - \nu_{S'}^z\|_{W}$ gives an upper bound for the SAS coefficient, $\beta.$ Such a bound is obtained from the straightforward use of an appropriate perturbation result on uniformly ergodic Markov chains.

\begin{theorem}
Given a uniformly ergodic Markov operator $\mathcal{P}^z_{\mu_S}$ constructed in Lemma \ref{lemma:markov}, and that the Lipschitz constant of $\nabla \ell(\cdot, w)$ on $\mathbb{R}^d$ is bounded above by $L_D$ for all $w \in \mathbb{W,}$ $\phi_S$ is SAS with stability coefficient 
$$
    \beta = {\cal O}\left( \frac{1}{n} \, \frac{L_D}{1 - \lambda}\right), 
$$
where $\lambda = \sup_z \lambda_z$ is the supremum over $z$ of the mixing rates of $\mathcal{P}^z_{\mu_S}$. 
\label{thm:perturbation_bound}
\end{theorem}

\begin{proof} (Sketch)
 The proof relies on the perturbation bounds of \citet{wasserstein}, Corollary 3.2. Note that when \eqref{eq:uniformErgodicity} is satisfied, and if a stochastic  perturbation $S'$ introduces a change $\delta \mathcal{P}^z_{\mu_S}$ to $\mathcal{P}^z_{\mu_S}$, the perturbation bound from \citep{wasserstein} gives 
\begin{align}
		\label{eqn:markovChainPerturbationBound}
		\left\|\nu_S^z - \nu_{S'}^z\right\|_{W} &\leq  
		 C\|\delta \mathcal{P}^z_{\mu_S}\|/(1-\lambda_z)
		,
\end{align}
where $\|\delta \mathcal{P}^z_{\mu_S}\|$ is the operator norm of $\delta \mathcal{P}^z_{\mu_S}$ induced by Wasserstein distance. From \eqref{eq:connectionWithSAS},
\begin{align}
  \label{eq:lossdiff}
    |\langle \ell_z\rangle_S - \langle \ell_z\rangle_{S'}| \leq    C\|\delta \mathcal{P}^z_{\mu_S}\|/(1 - \lambda_z) .
\end{align}
Since this holds for all $z,$ the right hand side of \eqref{eq:lossdiff} is an upper bound for the stability coefficient
 $\beta$ in Definition  \eqref{def:algorithmicStatisticalStability}, by taking the supremum over $z$. An upper bound on $\left\|\delta \mathcal{P}^z_{\mu_S}\right\|$ can be derived when the Lipschitz constant of $\nabla \ell(\cdot, w)$ is uniformly (in $w$) small on $\mathbb{R}^d.$  We discuss this further in Appendix \ref{sec:correlation}), where the size of $\delta \mathcal{P}^z_{\mu_S}$ is proved to be such that $\beta = {\cal O}(L_D/n).$ 
 \end{proof}

Theorem~\ref{thm:perturbation_bound} implies that \emph{an algorithm that exhibits faster convergence to the stationary measure on the loss space generalizes better}. The bound in Theorem~\ref{thm:perturbation_bound} implies that smaller $\lambda$ (faster convergence of the loss to the ergodic invariant measure) yields smaller upper bounds on $\beta$ (more statistically stable algorithm) and better generalization.
 
Strictly speaking, to determine $\lambda_z$, we must consider finite-dimensional approximations of the operator $\mathcal{P}_{\mu_S}^z$ (as done in the setting of chaotic systems in e.g., \citep{froyland, chekroun2014}) and compute its spectral decomposition. However, in our setting, it is difficult in practice to generate samples in the loss space that obey the Markov process defined in Lemma \ref{lemma:markov}. 
Instead, we argue for using the readily available non-Markovian loss process to estimate convergence to the equilibrium distribution. In particular, we hypothesize below that the auto-correlation function of the test loss can separate algorithms based on their SAS. 

Dropping the superscripts $z$, we define the auto-correlation $C_\ell(\tau)$ in a loss function $\ell$ (see also Remark \ref{rmk:auto-correlation} in Appendix \ref{sec:correlation}) by 
\begin{align}
   \label{eq:auto-correlation}
   C_\ell(\tau) := |\lim_{T\to\infty}\dfrac{1}{T}\sum_{t\leq T} \ell(w_t) \ell(w_{t+\tau}) - \langle\ell\rangle^2|/\langle\ell\rangle^2. 
\end{align}
That is, $C_\ell(\tau)$ gives the correlation between the random variables $\ell(w_t)/\langle \ell\rangle$ and $\ell(w_{t+\tau})/\langle \ell\rangle,$ where the randomness comes from the batch selection $\Xi_t$ and the initialization of weights in SGD, and only from the latter in GD. 


Suppose that the loss process is Markov, i.e., $\ell(w_{t+1})$ conditioned on $\ell(w_t)$ is independent of $\{\ell(w_{t'}): t' \leq t-1\}.$ 
Setting $t=0$ after a sufficiently long runup time, the function $C_\ell(\tau)$ in \eqref{eq:auto-correlation} is the auto-correlation function of a stationary Markov chain, which typically decays exponentially with $\tau.$ This decay rate is lower (slower correlation decay/longer correlation times) when $\lambda$ is closer to 1. Qualitatively, comparing two different stationary Markov chains, the one with higher values of $C_\ell$ has longer correlation times and hence, larger $\lambda.$

In general, however, the loss process  $\{\ell(w_t):t\geq 0\}$ is  non-Markovian. Thus, we do not expect $C_\ell(\tau)$ to decay with $\tau.$ This is because the correlation function $C_\ell$ encodes both the Markovian and non-Markovian components component of the loss dynamics (see Remark \ref{rmk:mori} in Appendix \ref{sec:correlation}; \citep{zwanzig2001nonequilibrium, KONDRASHOV201533}). Hence, we expect the correlation times in the loss process to not be equal to that of a Markov process generated according to $\mathcal{P}_{\mu_S}.$ For the purpose of qualitatively comparing the SAS coefficients of two algorithms, however, the loss process can be considered a heuristic for a Markov process generated according to $\mathcal{P}_{\mu_S}.$ Then, higher values of $C_\ell$ computed from the loss process indicate larger $\lambda,$ and hence less statistical stability.
 \begin{figure}
    \centering
    \includegraphics[width=0.3\textwidth]{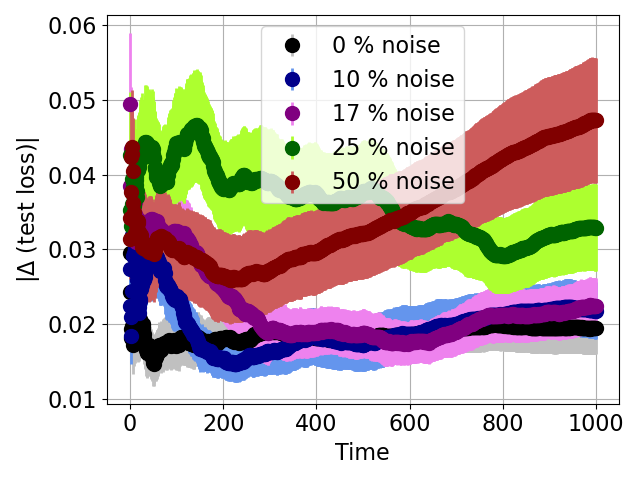}
    \includegraphics[width=0.3\textwidth]{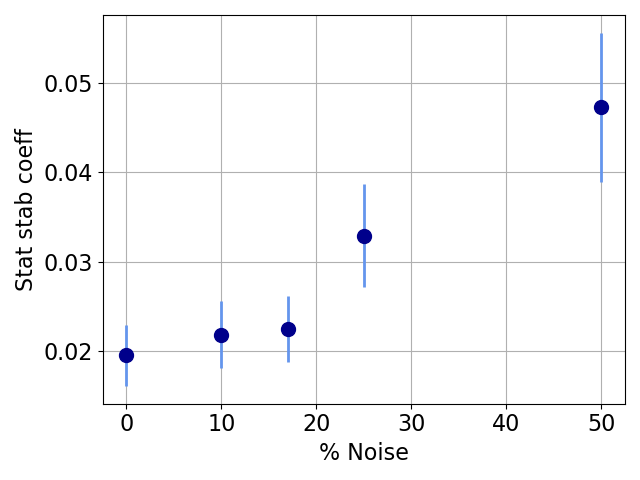}
    \includegraphics[width=0.3\textwidth]{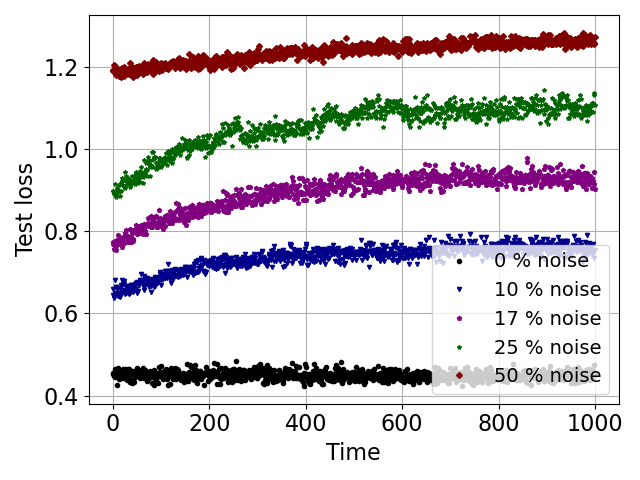}
    \caption{SAS computed from the VGG16 model. Left: The change in cumulative time averages of the test loss upon stochastic perturbation of the CIFAR10 dataset.  Center: Lower bound on the statistical stability coefficient $\beta$ (Definition \ref{def:algorithmicStatisticalStability}) computed using the test loss. Right: Test error timeseries.}
    \label{fig:statisticalStability}
\end{figure}
 \begin{figure}
    \centering
    \includegraphics[width=0.3\textwidth]{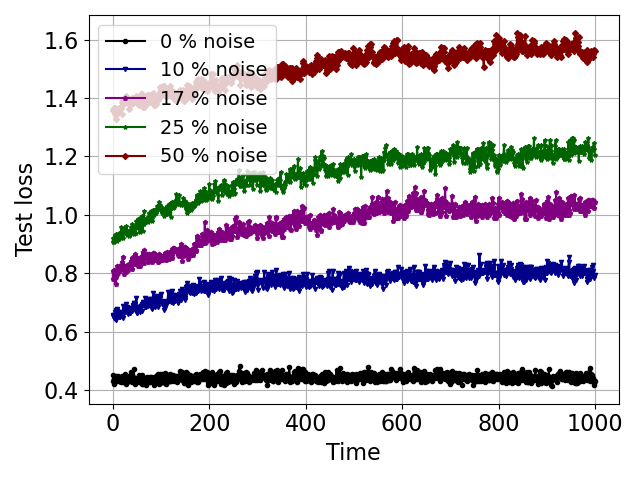}
    \includegraphics[width=0.3\textwidth]{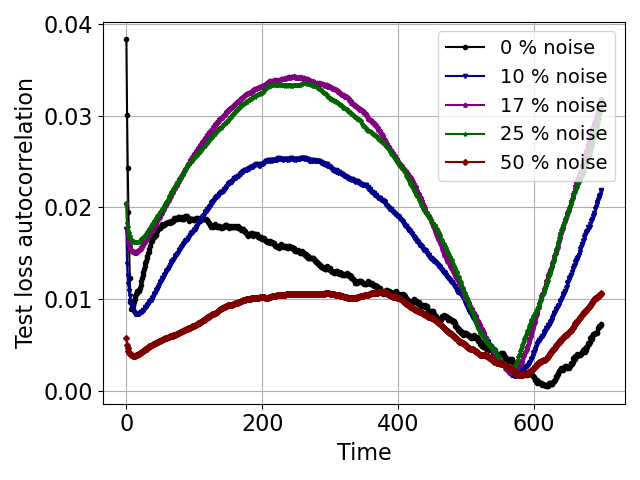}
     \includegraphics[width=0.3\textwidth]{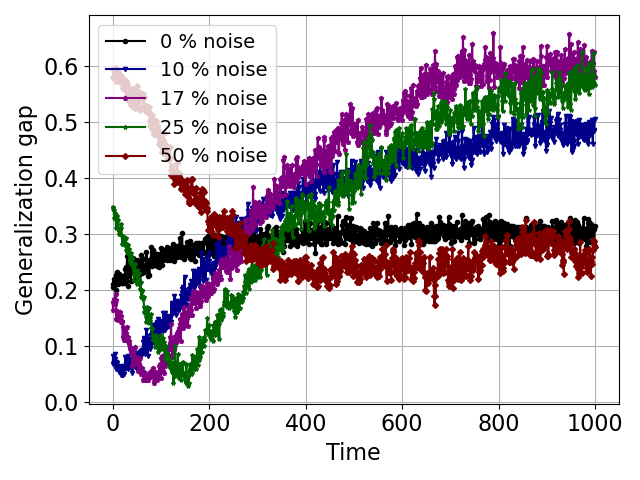}
    \caption{Predictors of generalization gaps on data with corrupted labels computed on the ResNet18 model. Left: Time series of the test loss. Center: Normalized auto-correlation of the test loss timeseries on the left.  The magnitude of the auto-correlation in the test loss is suggestive of the \textit{generalization gap} (right). For large percentage of noise, the gap decreases because the neural network cannot fit the training data well. 
    }
    \label{fig:generalization}
\end{figure}

 \section{Numerical results} 
\label{sec:numeric}
 We numerically validate the main ideas of section \ref{sec:stochasticStability} and \ref{sec:predict} on  VGG16 and ResNet18 models trained on the CIFAR10 dataset (see Appendix \ref{sec:resnet} for further numerical results). For all our experiments, $\phi_S$ is an SGD update with momentum 0.9, fixed learning rate 0.01 and batch size of 128. In all figures, ``time'' indicates number of epochs. 
 We generate different versions of the training set $S_p$ by corrupting CIFAR10's labels with probability $p,$ with $S_0$ being the original CIFAR10 dataset. Figures  \ref{fig:statisticalStability} and \ref{fig:generalization} show results corresponding to $p = 0, 0.1, 0.17, 0.25$ and $0.5.$ 
Each line in Figure \ref{fig:generalization} is a sample mean over 10 random initializations. 
 
In Figure \ref{fig:statisticalStability}, we numerically estimate a lower bound  on the SAS coefficient $\beta$ using its definition (see Definition \ref{def:algorithmicStatisticalStability}).  On the left, we plot the difference in test loss time-averages between between random orbits of $\phi_{S_p}$ and $\phi_{S'_p},$ with $S'_p$ being a stochastic perturbation of $S_p.$ The mean over 45 pairs of orbits is shown as dots and the error bars indicate the standard error in mean. The cumulative time average along orbits of length 1200 epochs are used as estimators for statistics. With these estimators for statistics, in Figure \ref{fig:statisticalStability}(center), we compute an estimate of the stability coefficient as the difference of the (estimated) test loss statistics at different values of $p.$ This is hence an estimate on the lower bound of $\beta,$ and clearly increases with $p.$ This illustrates that cases with worse generalization errors (Figure \ref{fig:statisticalStability} (right)) have larger lower bounds on $\beta.$ 

In Figure \ref{fig:generalization} (left), we show the test loss timeseries obtained with the ResNet18 model. Again, greater the noise corruption $p,$ the larger is the generalization error (estimated by test error), consistent with previous studies~\cite{zhang2021understanding,loukas2021what}). On the other hand, the generalization gap -- difference between training and test error -- is bigger for intermediate levels of noise and decreases when $p=0.5$, as shown in Figure \ref{fig:generalization} (center). The gap does not always increase with $p$ because, when the labels are close to random, the network cannot fit the training data and thus the training error is also large.

In Figure \ref{fig:generalization} (right), we plot $C_\ell(\tau)$ (Section \ref{sec:predict}) where $\ell$ is taken as the test loss. The test loss statistic $\langle \ell\rangle$ is estimated as a time average over 1200 epochs. At each $p,$ we show the sample average of the auto-correlations over 10 independent runs. 
We see that the magnitude of the test loss auto-correlations preserves the same order (across $p'$s) as the generalization gap (absolute difference in test and training losses) shown in Figure \ref{fig:generalization}(center). Hence, we empirically observe that the loss process codifies the phenomenological explanation for SAS that is expressed in \eqref{eq:lossdiff}.

\section{Discussion and conclusion}
Predicting generalization is an active area of research and has implications for more reliable use of machine learning. 
In this work, we introduce statistical algorithmic stability (SAS) and show how it implies generalization bounds. Here, we add a few important remarks. 

\emph{Statistical stability only requires robustness of statistics on loss space.} A central challenge in the analyses of non-convergent algorithms is the fact that there may be multiple, very different invariant measures $\mu_S$. For this reason, our stability criterion focuses on statistics of the loss function. This way, an algorithm can be stable even if the measures $\mu_S$ and $\mu_{S'}$ are not close in the total variation norm. In fact, an algorithm can be stable even if $\mu_S$ and $\mu_{S'}$ are mutually singular, if the loss functions have similar statistics. Since our SAS analysis hinges on a reasonable yet nonrestrictive assumption on ergodic properties of the algorithm, we believe it is broadly applicable.

\emph{Algorithms that converge are special cases of the above analysis.}
In Appendix \ref{sec:ntk}, we show that the proposed relationship (section \ref{sec:predict}) between statistical stability and convergence rates to stationary measures can also be observed in the Neural Tangent Kernel (NTK) regime \citep{jacot}. In this case, the training dynamics, $\phi_S,$ can be approximated by a linear function of the weights, which converges to a fixed point. Thus, this provides an alternative, ergodic theoretic, interpretation of generalization in the NTK regime \citep{aroraNTK, montanari2020interpolation, bartlett2021deep}.

\emph{Broader view.} While our focus is on algorithmic stability-based generalization, this work gathers more evidence to support the broader view \citep{wojtowytsch, suvrit} that exploiting theoretically and empirically available dynamical information about the training algorithm is a fruitful complement to understanding generalization from the optimization landscape and learning theory perspectives.

\vspace{1in}
{\bf Funding:} This work was partially funded by the NSF AI Institute TILOS, NSF award 2134108, and ONR grant N00014-20-1-2023 (MURI ML-SCOPE).

{\bf Acknolwedgments:} N.C. would like to thank Nandhini Chandramoorthy and Derek Lim for helping with the experimental setup and Sven Wang, Benjamin Zhang, Matt Li and Youssef Marzouk for valuable discussions. The authors also thank the reviewers for their constructive suggestions.


\clearpage
\appendix
\appendix

\section{Bifurcation analysis of smooth and non-convex optimization}
\label{sec:exampleDynamicsNonConvex}
In this section, we discuss some examples of non-convex optimization in one dimension performed with gradient descent (GD). We illustrate that, with increasing learning rate, the asymptotic behavior of orbits may alter from being periodic to quasiperiodic to chaotic. These qualitative changes are brought about through period-doubling  \emph{bifurcations}, which are observed in many physical systems (e.g., \citep{periodiDoubling1, periodicDoubling2, zhao2004observation}).  We consider smooth and non-convex objective functions of the form $\ell_s = g_s \circ g_s \circ g_s,$ with $g_s$ being the canonical quadratic map $g_s(w) := 1 - s w(1-w)$ of the unit interval $[0,1].$  Since $g_s$ is smooth, the compositions of $g_s$ with itself are smooth. 
The first bifurcation point appears just above $\eta = 2/|d^2 \ell_s/dw^2|,$ which is the stability threshold for convex optimization \citep{nesterov2003introductory}.

\begin{figure}[h]
    \centering
    \includegraphics[width=0.3\textwidth]{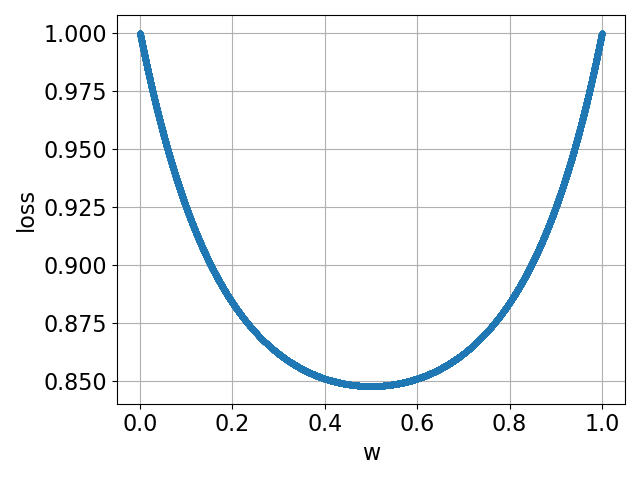}
    \includegraphics[width=0.3\textwidth]{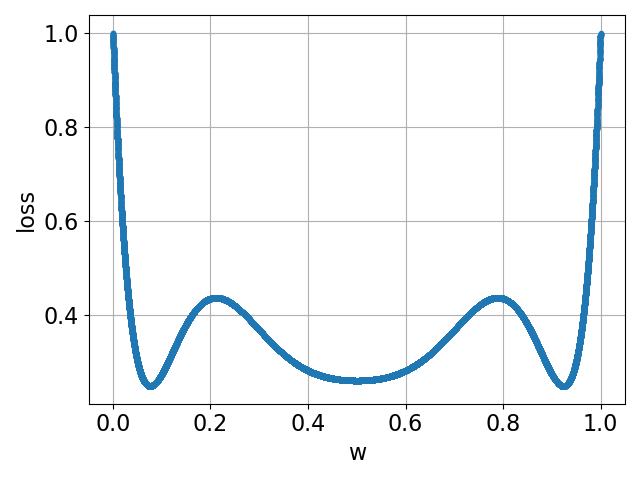}
    \includegraphics[width=0.3\textwidth]{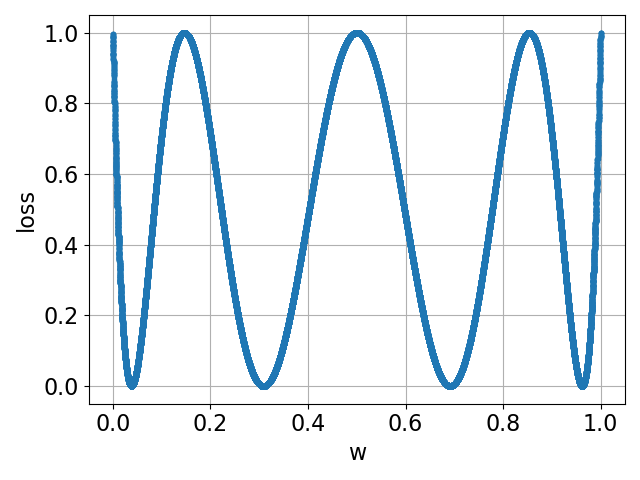} \\
    \includegraphics[width=0.3\textwidth]{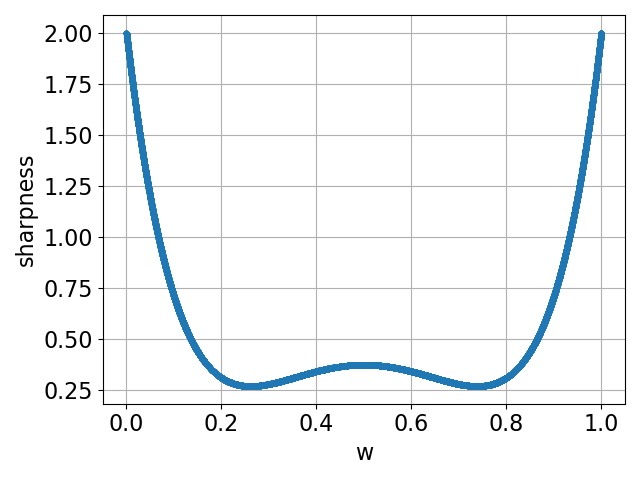}
    \includegraphics[width=0.3\textwidth]{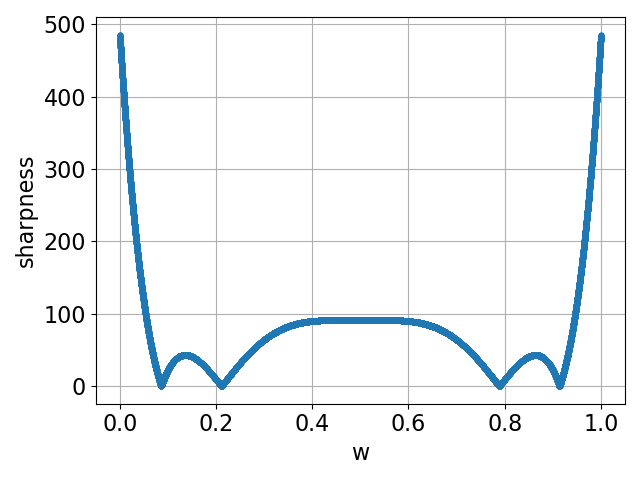}
    \includegraphics[width=0.3\textwidth]{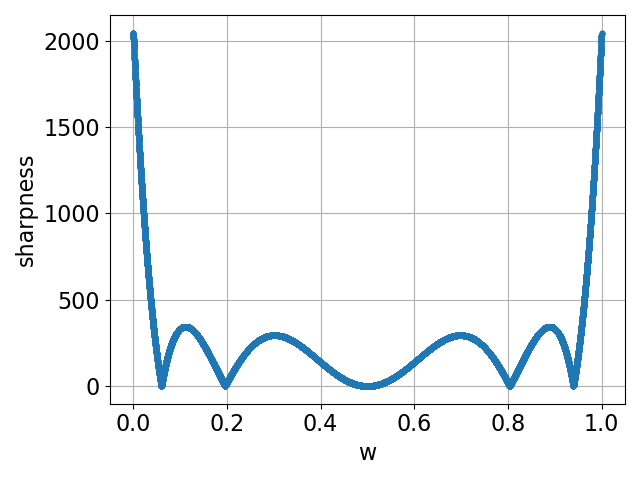} \\
    \includegraphics[width=0.3\textwidth]{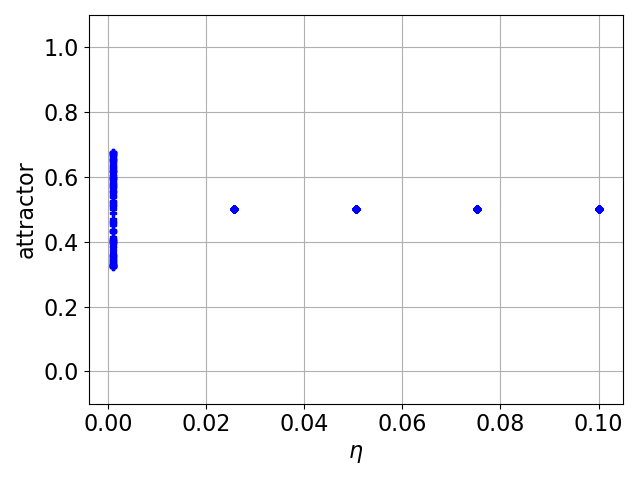}
    \includegraphics[width=0.3\textwidth]{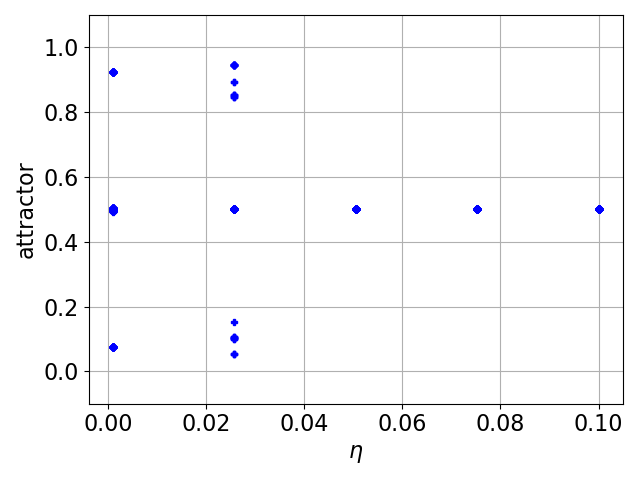}
    \includegraphics[width=0.3\textwidth]{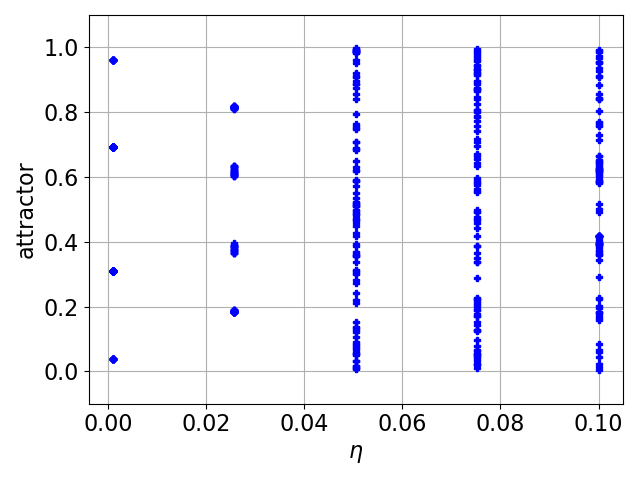} \\
    \includegraphics[width=0.3\textwidth]{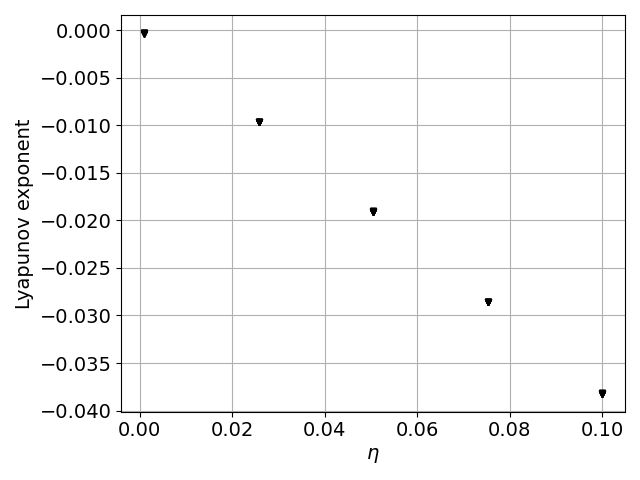}
    \includegraphics[width=0.3\textwidth]{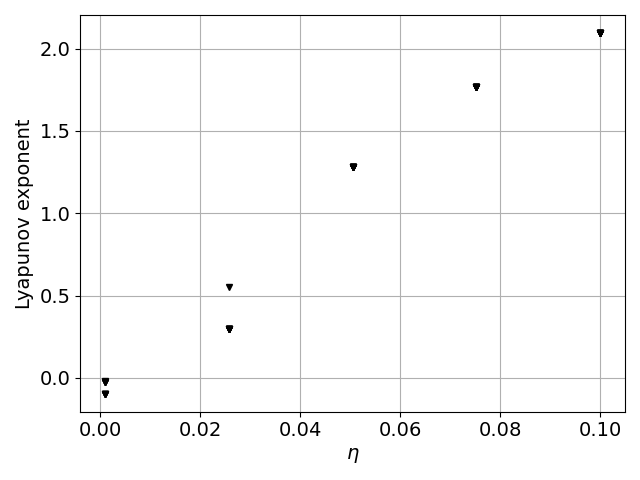}
    \includegraphics[width=0.3\textwidth]{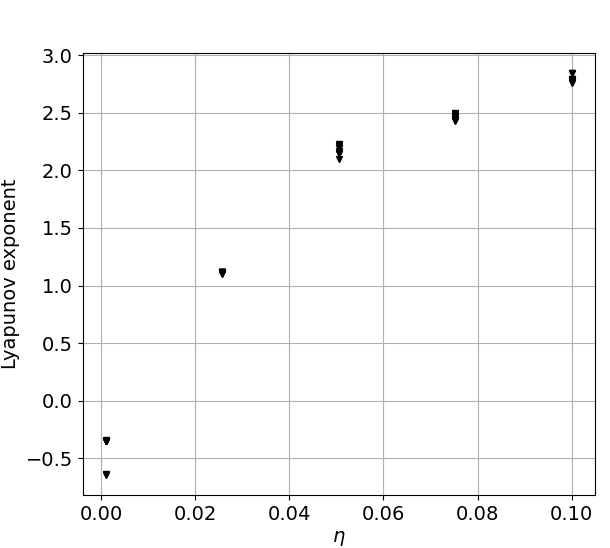}
    \caption{Bifurcation diagram of the toy model in section \ref{sec:exampleDynamicsNonConvex}. The left column corresponds to the dynamics $\phi_s$ (\eqref{eq:phis}) at $s=1,$ the center column at $s=3$ and the right column at $s=4.$ First row: The loss function $\ell_s$. Second row: Sharpness defined in section \ref{sec:exampleDynamicsNonConvex}. Third row: attractors starting from multiple different initial conditions, which turn out to be fixed points at small learning rates and periodic, quasiperiodic and chaotic orbits at larger learning rates ($\gg 2/$ sharpness). Fourth row:  Lyapunov exponent computed along different orbits showing chaos at large learning rates. 
    }
    \label{fig:bifurcationNonConvex}
\end{figure}

In Figure \ref{fig:bifurcationNonConvex}, the left, center and right columns correspond to $s=1,3$ and $4$ respectively. In the first row, we plot the loss functions $\ell_s$, which are non-convex (with multiple global minima) at $s = 3$ and $s=4.$ The second row shows the \emph{sharpness} -- absolute value of the second derivative, $a(w) = |d^2\ell_s(w)/dw|.$ At $s=1,$ the unique global minimum at $w = 0.5$ is flatter than the global minima at $s=3$ and $s=4.$  

The third row of Figure \ref{fig:bifurcationNonConvex} is a \emph{bifurcation} diagram, which shows the \emph{attractor} on the $y$-axis as a function of the learning rate. The attractor is approximated by the asymptotic orbits of the dynamics at multiple (100) different initial conditions chosen randomly on the unit interval. Note that at small values of $\eta < 2/\|a\|$ with $ \|a\|:=\sup_{w \in [0,1]} a(w)$ orbits from different initial conditions converge to fixed points corresponding to the local/global minima at each value of $s$. Periodic orbits emerge at $\eta > 2/\|a\|,$ that are ultimately shown to become chaotic for larger learning rates. This can be noted from the last row of Figure \ref{fig:bifurcationNonConvex}, where Lyapunov exponents (see e.g., \citep{katok, wilkinson2017lyapunov}) computed at different initial conditions are plotted. Given the gradient descent dynamics, 
\begin{align}
\label{eq:phis}
    \phi_s(w) = w - \eta \ell_s'(w),
\end{align}
where $f'(w) = (df/dw)(w),$ the
\emph{Lyapunov exponent}, $\lambda_s:[0,1]\to \mathbb{R}$ is defined as,
\begin{align*}
    \lambda_s(w)= \lim_{T\to\infty}\dfrac{1}{T}\sum_{t=0}^{T-1} \log|\phi_s'(\phi_s^t w)|.
\end{align*}
Roughly speaking, this function $\lambda_s(w)$ measures the asymptotic stability of infinitesimal linear perturbations along the orbit of $w.$ Here, we use the exponential notation, $\phi_s^t w = \phi_s \circ \phi_s^{t-1} w$ to denote compositions of $\phi_s.$ A positive Lyapunov exponent indicates dynamical instability, e.g., chaotic orbits. We see from the bottom row of Figure \ref{fig:bifurcationNonConvex} that in the range of learning rates considered, chaos is observed (positive Lyapunov exponents starting from uniformly random initial conditions) in the case of sharper minima, for larger learning rates. 
In ergodic systems, $\lambda_s(w)$ is independent of $w.$ Indeed, for larger learning rates, we see that $\lambda_s$ appears to be independent of the initial conditions (the bottom row of Figure \ref{fig:bifurcationNonConvex} shows $\lambda_s(w)$ for 100 different $w$). On the other hand, smaller learning rates where convergence to fixed points or periodic behavior is observed, the Lyapunov exponent converges to different negative values depending on the initial condition. This example supports the heuristic explanation for the primary assumption made in the main text about the asymptotic dynamics of learning algorithms. The assumption of lack of ergodicity ensures that the analysis is applicable to a range of learning rates, even those smaller than $2/\|a\|.$ 

\section{Proof of Theorem 1}
\label{sec:stabilityImpliesGeneralization}
In this section, we prove Theorem 1 from the main text. This result says that the statistical stability of an algorithm implies generalization.
\begin{figure}
    \centering
    \includegraphics[width=0.5\textwidth]{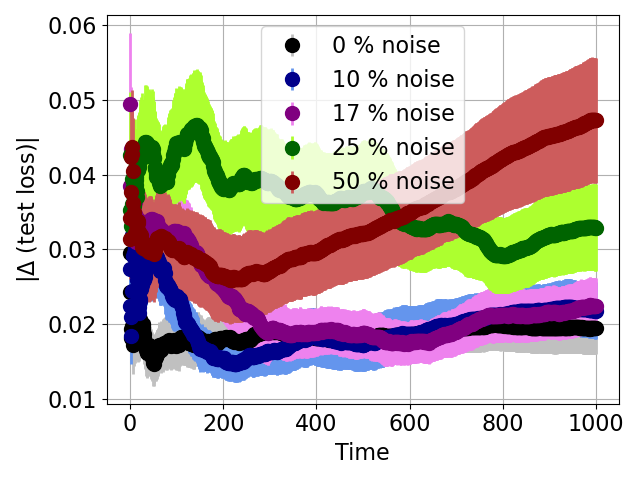}
    \caption{Absolute difference in time-averaged test loss due to stochastic perturbation computed with a ResNet18 architecture. The mean over 45 pairs of stochastic perturbations along with the standard error in mean are shown. }
    \label{fig:stabilityResNet}
\end{figure}
Recall that here, the population risk is defined as
\begin{align*}
		R_S = \Exp_{z\sim \mathcal{D}} \langle \ell_z\rangle_S.
\end{align*}
and the empirical risk is 
\begin{align*}
		\hat{R}_S = \dfrac{1}{n} \sum_{i=1}^n \langle \ell_{z_i}\rangle_{S}\end{align*}
We closely follow the proof strategies of \citet{bousquet2002stability} (see also \citep{algstab}). Define a function $\Phi_S := R_S - \hat{R}_S,$ whose expected value is
\begin{align}
		\notag
		\Exp_{S \sim \mathcal{D}^n} [\Phi_S] 
		&= \Exp_{S\sim \mathcal{D}^n} [R_S] - \Exp_{S \sim \mathcal{D}^n} [\hat{R}_S] \\
	\label{eqn:expandPhis}
		&= \Exp_{S \sim \mathcal{D}^n} \Exp_{z \sim \mathcal{D}} \langle \ell_z\rangle_S - \dfrac{1}{n}\sum_{i=1}^n \Exp_{S \sim \mathcal{D}^n} \langle \ell_{z_i}\rangle_S
\end{align}
Examining the second term, since $z_i$'s are chosen i.i.d. according to $\mathcal{D}$, $\Exp_{S \sim \mathcal{D}^n} \langle \ell_{z_i}\rangle_S$ is constant across $i$ and equal to $\Exp_{S \sim \mathcal{D}^n}  \langle \ell_{z_1}\rangle_S,$ where $z_1 \in S.$ As in the main text, let $S'$ denote any set that has at most one element different from $S, $ i.e., a stochastic perturbation of $S.$ Using the stochastic perturbation $S',$ we can rewrite $\Exp_{S \sim \mathcal{D}^n}  \langle \ell_{z_1}\rangle_S$ as $\Exp_{S \sim \mathcal{D}^n} \E_{z \in \mathcal{D}}  \langle \ell_{z}\rangle_{S'}.$  
Substituting this equivalent expression in \eqref{eqn:expandPhis}, and using the fact that $\phi_S$ is SAS with stability coefficient $\beta$: 
\begin{align}
	\notag	\left|\Exp_{S \sim \mathcal{D}^n} [\Phi_S]\right| &\leq \Exp_{S \sim \mathcal{D}^n} \Exp_{z \sim \mathcal{D}} \left|
		\langle	\ell_z\rangle_S - \langle \ell_{z}\rangle_{S'}\right| \\
	\label{eq:expPhis}
		&\leq \Exp_{S \sim \mathcal{D}^n} \sup_{z \in \mathbb{R}^{d}\times \mathbb{R}} \left|
		\langle	\ell_z\rangle_S - \langle \ell_{z}\rangle_{S'}\right| \leq \beta.
\end{align}
The proof is based on applying Mcdiarmid's inequality (see e.g., \citet{algstab}) to $\Phi_S$, which gives a high-probability upper bound on $\Phi_S - \E(\Phi_S)$ in terms of the deviation of $\Phi_S$ from a $\Phi_{S'}$. In order to obtain an upper bound for the latter quantity, note that
\begin{align}
\label{eq:phidiff}
		|\Phi_S - \Phi_{S'}| &\leq |R_S - R_{S'}| + |\hat{R}_S - \hat{R}_{S'}|.
\end{align}
Considering the difference of the population risks,
\begin{align}
\label{eq:gendiff}
	|R_S - R_{S'}| \leq  \Exp_{z\sim \mathcal{D}}\left|\langle \ell_{z}\rangle_S - \langle \ell_z\rangle_{S'}\right| \leq \beta. 	
\end{align}
Next considering the difference of empirical risks,
\begin{align}
\label{eq:empdiff}
	|\hat{R}_S - \hat{R}_{S'}| &\leq \dfrac{1}{n} \Big( \sum_{z_i \in S \cap S'}\left| \langle\ell_{z_i}\rangle_S - \langle \ell_{z_i}\rangle_{S'}\right| 
		+
 \left| \langle \ell_{z_k}\rangle_S
		\right| + |\langle\ell_{z'_k}\rangle_{S'}| \Big) \\
		&\leq \dfrac{1}{n}\left( (n-1)\beta +   2 L\right),
\end{align}
where, recall that $L := \sup_{z \in \mathbb{R}^d\times \mathbb{R}} \sup_{w\in \mathbb{W}} |\ell(z,\cdot)|.$ Putting \eqref{eq:gendiff} together with \eqref{eq:empdiff} into Inequality \eqref{eq:phidiff}, we obtain
$\left|\Phi_S - \Phi_{S'}\right| \leq 2\beta + \dfrac{1}{n} \left( 2 L - \beta\right).$
\begin{lemma}{(Mcdiarmid's inequality)}
Let $Z_1,\cdots,Z_n$ be random variables taking values in $\mathcal{Z}.$ A function $f:\mathcal{Z}^n\to \mathbb{R}$ is said to satisfy the bounded differences property if there exists a constant $c > 0$ such that
\begin{align*}
   \sup_{(z_1,\cdots,z_i, z_i',z_{i+1},\cdots,z_n)} |f(z_1,\cdots, z_n) - f(z_1,\cdots,z_{i-1} z_i',z_{i+1},\cdots, z_n)| < c,
\end{align*}
for all single coordinate changes. For an $f$ that satisfies the bounded differences with a constant $c > 0,$ given a $\delta > 0,$ with probability at least $1 - \delta,$
\begin{align*}
    \left|f(Z_1,\cdots,Z_n) - \E[f(Z_1,\cdots,Z_n)]\right| \leq c\sqrt{\dfrac{n}{2} \log(2/\delta) }
\end{align*}
\end{lemma}
Applying Mcdiarmid's inequality to $\Phi_S,$ and recalling \eqref{eq:expPhis}, we obtain the generalization bound  stated in Theorem 1 of the main text. That is, with probability $\geq 1- \delta,$
\begin{align}
    |\Phi_S - \E[\Phi_S]| = |R_S - \hat{R}_S - \E[\Phi_S]| \leq (2 \beta + \dfrac{1}{n}(2L - \beta)) \sqrt{\dfrac{n}{2} \log(2/\delta)}.
\end{align}
The above proof may also be repeated using the concentration inequality obtained in Theorem 4 of \citet{algstab}. This leads to a tighter bound analogous to Corollary 8 of \citet{algstab}.
\section{Predicting generalization with loss timeseries}
\label{sec:correlation}
In this section, we substantiate the connection drawn in section 4 of the main text between the rate of decay of correlations in the loss function and SAS. The study of dynamics lifted to the space of observables \citep{koopman} has a substantial precedent in dynamical systems theory (e.g., \citep{froyland, keller1999stability, dellnitz2000isolated}) and computational methods \citet{arbabi2017ergodic,mezic,williams2014kernel,korda2018convergence}. In particular, the idea of relating the correlation decay rate or the convergence rate of Fokker-Planck/Frobenius Perron operators with some notion of global sensitivity has also extensively appeared both in the statistical learning theory and SDE literature (see e.g.  \citep{bartlett2021deep, sirignano2022mean}) and in the dynamical systems literature (e.g.,  \citep{kato2013perturbation, chekroun2014}). In this work, inspired by the existence of such relationships in various contexts, we are able to show that the autocorrelations in the loss function can serve as predictors for the generalization gap. 

This section provides the complete setting for the results in section 4 of the main text and completes the proof of Theorem 2. First we define the transition operator for probability distributions on the weight space $\mathbb{W}$ and analogous operator on the loss space.

\textbf{Markov operator for weight space.} Recall that $\Xi_t$ is a batch of $m$ indices chosen uniformly from the set $[n] := \left\{ 1, \cdots, n\right\}.$ For GD, $\Xi_t$ is deterministic and equal to the set $[n]$. Let $(\Omega, \Sigma,\mathbb{P})$ be a probability space. Let $\mathcal{K}_S:\mathbb{W}\times \mathcal{B}(\mathbb{W})\to \mathbb{R}^+$ be the Markov kernel associated with the update $\phi_S,$ i.e., for a Borel subset $A \in \mathcal{B}(\mathbb{W})$ and a point $w \in \mathbb{W},$ 
\begin{align}
\label{eq:markovKernel}
    \mathcal{K}_S(w, A) = \E_{\Xi_t} \mathbb{P}(\phi_S(w) \in A|\Xi_t), 
\end{align}
where $\mathbb{P}(\phi_S(w) \in A|\Xi_t)$ is the probability of the event that $\phi_S(w) \in A$ when the $m$ indices $\Xi_t$ are chosen. Correspondingly, we may define the Markov operator $\mathcal{P}_S,$ also called the Frobenius-Perron operator \citet{lasota1998chaos},  on the space of probability measures on $\mathbb{W},$
\begin{align}
\label{eq:frobeniusPerron}
    \mathcal{P}_S \mu (A) = \E_{w \sim \mu}[\mathcal{K}_S(w, A)].
\end{align}
 From the above definition, it is clear that any $\phi_S$-invariant probability measure $\mu$ is an eigendistribution of $\mathcal{P}_S$ with eigenvalue 1. In our setting (see section 2 of the main text), there are potentially multiple eigendistributions corresponding to eigenvalue 1. Each invariant measure $\mu_S$ also defines different transition probabilities on the weight space $\mathbb{W}$: 
 \begin{align}
     P_{\mu_S}(A, B) = \dfrac{\E_{\mu_S}[\mathbbm{1}_A \mathcal{K}_S(\cdot, B)]}{\mu_S(A)}.
 \end{align}
\textbf{Markov operator for the loss.} 
 Now, instead of the phase space $\mathbb{W},$ for each $z,$ consider the image, $I_z \equiv \ell(z, \mathbb{W}) \subseteq [0, L]$ of $\mathbb{W}$ under $\ell(z,\cdot),$ with $L$ as defined in Theorem 1. 
 Let $\mathcal{B}(I_z)$ denote the Borel sigma algebra on $I_z.$ Analogous to the kernel \eqref{eq:markovKernel}, we may now define a kernel $\mathcal{K}_{\mu_S}^z,$ for an $(\xi, E) \in I_z \times \mathcal{B}(I_z),$ as 
 \begin{align}
     \label{eq:lossKernel}
 \mathcal{K}_{\mu_S}^z(\xi, E) = P_{\mu_S}(\ell(z, \cdot)^{-1}(\xi),\ell(z, \cdot)^{-1}(E)). 
 \end{align}
  This, in turn, gives rise to a Markov operator analogous to the Frobenius-Perron operator \eqref{eq:frobeniusPerron} on the full-dimensional weight space:
 \begin{align}
     \label{eq:lossFrobeniusPerron}
     \mathcal{P}_{\mu_S}^z\nu(E) = \E_{\xi\sim \nu}[\mathcal{K}_{\mu_S}^z(\xi, E)].
 \end{align}
 Note that this operator is well-defined when the level sets of $\ell_z$ are $\mu_S$-measurable. One sufficient condition for this is when the foliation of $\mathbb{W}$ by these level sets is subordinate to a measurable partition.
 We may then consider disintegrations of $\mu_S$ on this measurable partition and define the kernel \eqref{eq:lossKernel} using conditional measures supported on elements of the partition. Further, it is clear that $\mathcal{P}_{\mu_S}^z$ satisfies the properties of a Markov operator (positive unity preserving contraction).
 
 Thus, in order to prove Lemma 1, it remains to show that the operator defined by \eqref{eq:lossFrobeniusPerron} is mixing. For this, we make an additional assumption. 
 We assume that the Frobenius-Perron operator $\mathcal{P}_S$ mixes to the measure $\mu_S$ starting from any measure of the form $\mu = \ell(z,\cdot)^{-1}\nu,$ for an absolutely continuous probability measure $\nu$ on $[0,L].$
That is, 
\begin{align}
\label{eq:mixing}
   \| \mathcal{P}_S^t \mu_S - \mathcal{P}_S^t \mu \|_{\rm TV} =  {\cal O}(\zeta^t \|\mu_S - \mu\|_{\rm TV}),
\end{align}
where $\|\cdot - *\|_{\rm TV}$ indicates the total variation distance and $\zeta \in (0,1)$ is the rate of mixing. On the other hand, since $\mathcal{P}_S^t\mu \to \mu_S$ weakly, $\nu^z_t := \left(\mathcal{P}^z_{\mu_S}\right)^t \ell(z,\cdot)_\sharp \mu \to \nu_S$ weakly on $\mathcal{B}(I_z).$ Intuitively, we expect that rate of mixing of the latter, say $\lambda^z \in (0,1)$ correlates with $\zeta$ since
\begin{align}
\notag
\left|\langle \ell_z\rangle_S - \E_{\xi \sim \ell(z,\cdot)_\sharp \mu_t}[\xi]\right| &\leq \sup_{f, \|f\|\leq 1} \|\E_{w\sim \mu_S}[f(w)] - \E_{w \in \mu_t}[f(w)] \|  \\
\label{eq:vaguebound}
&= \|\mu_S - \mu_t\|_{\rm TV}.
\end{align}
Assuming that $\mathbb{W}$ is a Polish space, the above relationship \eqref{eq:vaguebound} conveys that when a measure $\mu_t$ converges to $\mu_S$ in the TV norm, expectations with respect to $\mu_t$ of all continuous functions, of which $\left\{\ell(z, \cdot)\right\}_z$ is a subset, also converge to expectations with respect to $\mu_S.$ Thus, intuitively we expect that when $\sup_z\lambda^z$ is a lower bound for $\zeta$ (\eqref{eq:mixing}). In the main text, below Lemma 1, we state the uniform ergodicity of $\mathcal{P}_{\mu_S}^z$ in terms of the Wasserstein norm. This holds from \eqref{eq:vaguebound} since convergence in TV distance implies convergence in Wasserstein. This concludes the proof of Lemma 1. Finally, note that our setting is different from previous works (\citep{chekroun2014} and references therein) in the dynamics literature that use observable-specific Markov operators, in that we do not assume uniqueness of the ergodic, invariant measure on the full-dimensional (weight) space. 

\textbf{Effect of stochastic perturbations} In section 4 of the main text, we use the perturbation theory of mixing Markov operators to relate the SAS coefficient to the mixing rate of $\mathcal{P}_{\mu_S}^z.$ This perturbation bound (from \citep{wasserstein}) is given in terms of the perturbation $\delta \mathcal{P}_{\mu_S}^z$ to the operator $\mathcal{P}_{\mu_S}^z$ when a stochastic perturbation $S \to S'$ is applied to the training set. Here we discuss the size of $\delta \mathcal{P}_{\mu_S}^z,$ completing the proof of Theorem 2 in the main text. First note that $\|\delta P_{\mu_S}^z\| \leq \|\mathcal{P}_S - \mathcal{P}_{S'}\|,$ and hence it suffices to obtain an upper bound for $\|\mathcal{P}_S - \mathcal{P}_{S'}\|.$ The perturbation to Markov kernel in the weight space due to a stochastic perturbation is given by
\begin{align}
\label{eq:KSdiff}
    \mathcal{K}_S(w, A) - \mathcal{K}_{S'}(w, A) = \sum_{\Xi \in \Delta} \Big(\mathbb{P}(\phi_S(w) \in A|\Xi) - 
    \mathbb{P}(\phi_{S'}(w) \in A|\Xi)\Big),
\end{align}
where $\Delta$ is the set of $m$ indices from $[n]$ that contain $k,$ the index at which $S$ and $S'$ differ. When $\Xi$ is a uniform random variable as we have assumed, the cardinality $\Delta$ is ${n-1 \choose m-1}/{n \choose m} = m/n.$ 
This leads to the following upper bound on the perturbation size in the Wasserstein norm,
\begin{align}
\notag
    \|\mathcal{P}_S\mu - \mathcal{P}_{S'}\mu \|_{\rm W} &:= \sup_{f, \|f\|_{\rm Lip}\leq 1} |\E_{w \sim \mathcal{P}_S\mu} f - \E_{w \sim \mathcal{P}_{S'}\mu} f| \\
    \notag
    &= \sup_{f, \|f\|_{\rm Lip}\leq 1} |\E_\Xi \E_{w \sim \mu} f\circ \phi_S - \E_\Xi \E_{w \sim \mu} f\circ\phi_{S'}| \leq \dfrac{m}{n}\sup_{w \in \mathbb{W}}
    \|\phi_S(w) - \phi_{S'}(w)\| \\
    \notag
    &\leq \eta \dfrac{m}{n} \sup_{w \in \mathbb{W}}
    \|\nabla L_S(w) - \nabla L_{S'}(w)\| \leq \dfrac{m\eta}{n^2} \sup_{w}\|\nabla\ell(z_k, w) - \nabla\ell(z_{k'}, w)\| \\
    &\leq c\: m\: L_D \eta/n^2.
\end{align}
In order to apply this bound, we use a result from the perturbation theory of Markov chains from \citep{wasserstein} (Corollary 3.2 in the asymptotic limit) as explained in the proof sketch of Theorem 2 in section 4. Thus, we obtain again, an upper bound for $\beta = {\cal O}(L_D/(1-\lambda)n)$ as claimed in Theorem 2.

\begin{remark}
\label{rmk:banach}
In this section as well as section \ref{sec:predict}, we use inequalities between different norms on the space of finite signed measures. This is a Banach space with the total variation norm, isomorphic to $L^1(\Lambda)$ for some background measure $\Lambda.$
The inequalities (for instance, \eqref{eq:connectionWithSAS}) follow from the dual characterization of the norms. 
\end{remark}
\begin{remark}
\label{rmk:auto-correlation}
Although the definition of the autocorrelation in section 4 appears to be a function of $w_0,$ $C_\ell(\tau)$ does not depend on $w_0$ due to Assumption \ref{ass:ergodicity}. Note that the rate of convergence to equilibrium, the coefficient $\lambda$ in the definition of uniform ergodicity, also determines the correlation decay rate when $\ell$ is initialized out of equilibrium (That is, $\ell(w_0)$ does not sample $\nu_S$). 
\end{remark}
\begin{remark}
\label{rmk:mori}
{\rm 
One way to understand the non-Markovian loss process is through the Mori-Zwanzig formalism \citep{zwanzig2001nonequilibrium} that originated in statistical mechanics and has found extensive applications in deriving reduced-order models for complex physics (e.g. see \citep{LIN2021109864, KONDRASHOV201533}). In this formalism, we can consider the exact evolution of a finite set of observables, $\Psi(w) = [\psi_1(w), \cdots, \psi_p(w)]$ such that $\ell \in {\rm span}\{\Psi\}$ (here, $\ell := \ell_z$ for an arbitrary $z$). At time $t,$ $\Psi_t := \Psi(w_t)$ can be written as a sum of three terms: 1) a Markov term that depends on the values $\Psi_{t-1},$ 2) a non-Markovian memory term that is a function of $\{\Psi_{t-k}: t \geq k > 1\},$ and 3) a noise term that is a function of $w_0.$ The first two terms depend on the dynamics $\phi_S$ and are hence different for different values of $S.$ The autocorrelation function $C_\ell$ has contributions from both the Markovian and non-Markovian terms.
}
\end{remark}

\section{Stability experiments on ResNets}
\label{sec:resnet}          
We obtain similar results for SAS with the ResNet18 model as with the VGG16 model shown in Figure 3 of the main text. In Figure \ref{fig:stabilityResNet}, we plot the difference in the cumulative average of the test loss at runs with the ResNet18 architecture. The difference is taken between two SGD runs with the same parameters as in section 5 of the main text and with training data that are stochastic perturbations of each other. We consider 45 pairs of stochastically perturbed datasets (see section 2 of main text for definition of stochastic perturbation) each for each value of $p$. The value of $p = 0, 0.1, 0.17, 0.25, 50$ indicates the probability of error injected into the labels of the CIFAR10 dataset. The mean of the absolute difference in the cumulative test loss is shown in dark colors while the standard error in mean in the corresponding lighter color. The time averages are calculated over 1200 epochs after a run up time of 200 epochs. The results indicate that greater the noise probability $p,$ greater the estimate of SAS (less statistical stability). Hence, these results indicate that statistical algorithmic stability (see section 2) correlates with the generalization performance of the models at the difference label corruption (noise) levels. Furthermore, considering together with the VGG16 results presented in the main paper, this relationship between SAS and generalization holds across the different architectures we have employed in this paper.  

The generalization gap plotted in Figure 3 (center) (of the main text) is the cumulative time average (ergodic average) of the absolute difference between the test and training errors. This is approximately equal to (strictly, an upper bound for) $|R_S - \hat{R}_S|.$ A loose upper bound for this quantity comes from the theoretical generalization bound (in Theorem 1). In practice, this quantity is estimated to be small relative to the test error with corrupt datasets. When the noise probability is 0\% (original dataset), the gap estimate is about 80\% of the test error because the training error is small. But, when the noise probability is 50\%, the gap estimate is about 20\% of the test error since the training error is also large. We may not observe this reduction in the generalization gap if the errors (test and training) were defined using pointwise values. That is, since, $\sup_w |E_{z \sim D} \ell_z(w) - (1/n)\sum_{i=1}^n \ell_{z_i}(w)|\geq |R_S - \hat{R}_S|,$ upon early stopping when training error is low, we may not observe this phenomenon. 

Finally, we remark that our empirically estimated autocorrelation function serves as a proxy until a more sophisticated method for the estimation of $\lambda$ is developed. As mentioned in the main text (section 4), this is a challenging problem that is beyond the scope of this work.

\section{Revisiting stability in the linear regime}
\label{sec:ntk}

Since its introduction by \citet{jacot}, training in the Neural Tangent Kernel (NTK) regime has been analyzed thoroughly in numerous works, wherein its convergence to kernel ridge regression has been formally proved in two-layer, infinitely wide networks \citep{montanari2020interpolation, bartlett2021deep}, infinitely wide fully connected networks \citep{arora2019fine}, convolutional networks \cite{aroraNTK} and so on. Using its equivalence with kernel ridge regression, the generalization properties under the NT regime have also been well-studied (see \cite{bartlett2021deep} for a review). Here, we revisit the NT regime with the purpose of demonstrating that the analysis in the present paper also holds when $\phi_S$ achieves a fixed point.

The Neural Tangent Kernel (NTK) model is an approximation of a neural network whose parameters remain close to initialization during training. Given that training in the NT regime is well-approximated by the dynamics of linear regression (see Theorem 3.1 and 3.2 of \citet{aroraNTK}), in order to apply our dynamics-based generalization analyses to the NTK regime, we need only consider linear regression dynamics. That is, let $\tilde{w}_t$ be an orbit of a perturbed dynamics (the linear regression dynamics, see Lemma 1 of \citep{aroraNTK}) close to an orbit $w_t = \phi_S(w_{t-1})$ for all time, so that, for some small $\epsilon > 0,$
\begin{align}
    \lim_{t\to\infty}\|w_t - \tilde{w}_t\| = \|w^* - \tilde{w}^*\| \leq \epsilon.
\end{align}
As mentioned in the main text (section 3), SAS reduces to the standard notion of algorithmic stability (see e.g., \citep{mohri} chapter 14 for a survey) when the dynamics converges to a fixed point. Now suppose that the dynamics $\tilde{w}^*$ is algorithmically stable. That is, for all stochastic perturbations $S'$ of $S,$ and for all $z,$ 
\begin{align*}
    |\ell(\tilde{w}_S, z) - \ell(\tilde{w}_{S'}, z)| \leq \beta.
\end{align*}
Assuming Lipschitz loss with Lipschitz constant $C_{\rm Lip}^z$ and denoting by $C_{\rm Lip} = \sup_z C_{\rm Lip}^z$ 
\begin{align}
    \notag
    |\ell(w_S^*, z) - \ell(w_{S'}^*, z)| &\leq |\ell(\tilde{w}_S^*, z) - \ell(\tilde{w}_{S'}^*, z)| + |\ell(w_S^*, z) - \ell(\tilde{w}_{S}^*, z)| \\
    &+ |\ell(w_{S'}^*, z) - \ell(\tilde{w}_{S'}^*, z)| \\
    &\leq \beta + 2 C_{\rm Lip}\epsilon.
\end{align}
That is, the NTK orbit is stable with the stability coefficient $\beta + 2 C_{\rm Lip}\epsilon.$ Thus, in order to prove the algorithmic stability of the NTK orbit, it is enough to consider the stability of the linear regression orbit, which is a linear dynamical system as we describe below. 

The second main idea that we develop is the prediction of the stability coefficient by the rate of decay of correlations. In this case, we show that the speed of convergence to the fixed point determines the generalization properties via algorithmic stability.
We derive stability-based generalization bounds, an alternative to Rademacher complexity-based bounds in \cite{arora2019fine}. Comparing with existing generalization results in the well-understood NT regime is an ideal test bed for the alternative dynamical perspective of this present work.

\textbf{Linear dynamics} Let $w_r\in \mathbb{W}$ be the parameter at which a NN $h_{\rm NN}(\cdot,w_r)$ is the zero function from $\mathbb{R}^d$ to $\mathbb{R},$ i.e., $h_{\rm NN}(x,w_r) = 0,$ for all $x.$  Now consider training the weights $w \in \mathbb{W}$ of the NN, with the initialization $w_0 = w_r.$ With GD on the squared loss $L_S(w) = (1/2)\sum_{i=1}^n (y_i - h_{\rm NN}(x_i, w))^2,$ and learning rate $\eta > 0,$ the dynamics of the weights are as follows,
\begin{align}
    \phi_S(w) 
    &= w + \eta \: \nabla h_{\rm NN}(X, w)^T (Y - h_{\rm NN}(X, w)). 
\end{align}
Here, $X = [x_1,\cdots,x_n]^T \in \mathbb{R}^{d\times n}$ and the notation $h_{\rm NN}(X,w) \in \mathbb{R}^n$ represents $[h_{\rm NN}(x_1,w),\cdots,h_{\rm NN}(x_n,w)]^T.$ Note that the above dynamics $\phi_S(w)$ is a nonlinear function of $w.$ Now we consider the NTK setting described in \cite{bartlett2021deep} so that we replace $h_{\rm NN}$ with its linearization about $w_r,$  $\nabla h_{\rm NN}(x, w_r)(w - w_r).$ With this linearization about $w_r,$ the above dynamics $\phi_S$ becomes linear in $w$, 
\begin{align}
    \tilde{\phi}_S(w) = w + \eta \Phi_S^T (Y_S - \Phi_S (w - w_r)),
\end{align}
where $Y_S = [y_1, \cdots, y_n]^T \in \mathbb{R}^n.$
Recalling that $\mathbb{W} \subset \mathbb{R}^{d_w},$ $\Phi_S$ is an $n\times d_w$ matrix with the $i$th row being $\nabla h_{\rm NN}(x_i, w_r).$ In the NTK regime, the dynamics $\phi_S$ is well-approximated by $\tilde{\phi}_S$ (see Theorem 5.1 of \citep{bartlett2021deep} for conditions under which the approximation holds). That is, the linear dynamics close to the NTK dynamics, referred to as the linear regression dynamics above, is the following,
\begin{align}
\label{eq:linearDynamics}
    \tilde{w}_{t+1} = A_S\tilde{w}_t + b_S, 
\end{align}
where $K_S := \Phi_S^T \Phi_S,$ $A_S := (I - \eta K_S) \in \mathbb{R}^{d_w\times d_w}$ and $b_S := \eta \left(K_S w_r + \Phi_S^T Y_S \right) \in \mathbb{R}^{d_w}.$ The dynamics converges to a fixed point $\tilde{w}^*_S = A_S \tilde{w}^*_S + b_S,$ as long as $\|A_S\| < 1.$

\textbf{Evolution on function space} Note that the invariant distribution of the above dynamics is singular: the delta distribution centered at $\tilde{w}_S^*.$ In order to repeat the analysis in section 4 for this special case, we need to obtain the relationship between the rate of decay of correlations with respect to this invariant measure and the stability of the fixed point. First isolating the rate of decay of correlations, this rate is equal to the second largest eigenvalue of the associated Frobenius-Perron on $L^2(\mathbb{W})$ or equivalently, the Koopman operator on $L^2(\mathbb{W})$ (since these two linear operators are adjoint to each other, they share isolated spectra). Since the dynamical system is linear, it is easy to verify that the eigenvalues of $A_S$ are also Koopman eigenvalues. We can also check that the eigenfunctions corresponding to eigenvalue $\theta_i$ are of the form $v_i^T w + 1/(\theta_i -1) v_i^T b_S,$ where $v_i$ are left eigenvectors of $A_S.$  

Note that $A_S$ is a symmetric matrix whose largest absolute eigenvalue is equal to $1 - \eta \theta_{\rm min},$ where $\theta_{\rm min} > 0$ is the smallest eigenvalue of the NTK $\Phi_S \Phi_S^T.$ 

\textbf{Stability of the fixed point} In order to relate the rate of convergence, $\lambda :=1 - \eta \theta_{\rm min}$ with SAS in this case, we now describe SAS in this regime. As we noted previously, SAS reduces to the algorithmic stability of the fixed point for the dynamics \eqref{eq:linearDynamics}. In order to deduce the stability of the fixed point to stochastic perturbations in the input, we need to recognize that the fixed point is an exact interpolant. 

 We can check that  $\tilde{\phi}_S(\tilde{w}_S^*) = \tilde{w}_S^*$ iff $Y_S = \tilde{\phi}_S(\tilde{w}_S^* - w_r).$ That is, the function $\nabla h_{\rm NN}(x,w_r)(w_S^* - w_r)$ exactly interpolates at the data points, among the class of linearized functions. In other words, the fixed point is $\tilde{w}_S^* = w_r + a_S,$ where $a_S$ is the minimum norm interpolation solution given by $a_S = \Phi_S^T (\Phi_S \Phi_S^T)^{-1} Y_S.$ 
 Thus, having a closed form expression for $w_S^*,$ we can obtain an upper bound on the stability of the algorithm $\phi_S.$

Since $\hat{K}_S := \Phi_S\Phi_S^T$ and its inverse are symmetric, $\|(\hat{K}_S)^{-1}\|$ is also the maximum eigenvalue of $(\hat{K}_S)^{-1}.$ A stochastic perturbation $S'$ of $S$ introduces a rank-one change denoted $\delta K$ to $(\hat{K}_S)^{-1}.$ From Weyl's inequality, $\|(\hat{K}_S)^{-1} - (\hat{K}_{S'})^{-1}\| \leq \|\delta K\|.$ In the case of Lipschitz loss, an upper bound on $\beta$ therefore depends on the maximum eigenvalue of $\hat{K}_S^{-1}$, which is equal to $1/\theta_{\min}.$ Thus, we see that a smaller  $\theta_{\min}$ implies a smaller rate of convergence (slower convergence) as well as a larger upper bound on $\beta$ (lesser algorithmic stability). Hence, the linear regime also supports the analysis in section 4, which discusses a more general scenario of convergence of weights in distribution.
\begin{remark}
As an aside that applies to the entire paper, we clarify that by ``statistics'' we refer to statistics over the parameter space $\mathbb{W}.$ The distribution over the weight space is specified in each context. Recall that the randomness in the sense of randomness  over the weights arises due to the stochastic nature of SGD as well as the randomness over initial conditions. 
\end{remark}
\begin{remark}
While we have considered linearization about a point, we may repeat the above analysis by considering another linear network via modifying the definition of the empirical matrix $\Phi_S\Phi_S^T$ to $K_S = E_{w \in \mathcal{D}_w} \nabla h_{\rm NN}(X,W)\nabla h_{\rm NN}^T(X,w).$ This is the well-studied limit of the empirical kernel as the number of neurons $N$ tends to infinity.  
\end{remark}

\bibliographystyle{abbrvnat}
\bibliography{main}

\end{document}